\documentclass[letterpaper]{article} % DO NOT CHANGE THIS

\usepackage{aaai24}  % DO NOT CHANGE THIS - camera-ready

\usepackage{times}  % DO NOT CHANGE THIS
\usepackage{helvet}  % DO NOT CHANGE THIS
\usepackage{courier}  % DO NOT CHANGE THIS
\usepackage[hyphens]{url}  % DO NOT CHANGE THIS
\usepackage{graphicx} % DO NOT CHANGE THIS
\usepackage{natbib}  % DO NOT CHANGE THIS AND DO NOT ADD ANY OPTIONS TO IT
\usepackage{caption} % DO NOT CHANGE THIS AND DO NOT ADD ANY OPTIONS TO IT
\usepackage{marvosym}
\usepackage{algorithm}
\usepackage{newfloat}
\usepackage{listings}
\DeclareCaptionStyle{ruled}{labelfont=normalfont,labelsep=colon,strut=off} % DO NOT CHANGE THIS
\floatstyle{ruled}
\newfloat{listing}{tb}{lst}{}
\floatname{listing}{Listing}
\usepackage[utf8]{inputenc} % allow utf-8 input
\usepackage[T1]{fontenc}    % use 8-bit T1 fonts
\usepackage{booktabs}       % professional-quality tables
\usepackage{amsfonts}       % blackboard math symbols
\usepackage{nicefrac}       % compact symbols for 1/2, etc.
\usepackage{microtype}      % microtypography
\usepackage{xcolor}         % colors
\usepackage{svg}

\usepackage{xspace}
\usepackage{multirow}
\usepackage{graphicx}
\usepackage{adjustbox}
\usepackage{makecell}
\usepackage{mathtools}
\usepackage{algpseudocode}
\usepackage{lscape}
\usepackage{rotating}

\usepackage{amsmath}
\usepackage{epigraph}

\AtBeginDocument{}

\newcommand{\ie}{{\it i.e.}}

\newcommand{\std}[1]{\tiny{$\pm$#1}}
\newcommand{\cell}[2][1]{\makecell{#1 \\ \std{#2}}}

\newcommand{\ocell}[2][1]{#1\std{#2}}

\usepackage{enumitem}
\usepackage{amsmath}
\usepackage{amssymb}
\usepackage{amsthm, thmtools}
\usepackage{multirow}
\usepackage{pifont} % cmark & xmark / http://ctan.org/pkg/pifont 
\usepackage{lipsum}

\newcommand{\cmark}{\ding{51}}
\newcommand{\xmark}{\ding{55}}

\definecolor{mblue}{rgb}{ 0,  .439,  .753}

\title{When Model Meets New Normals: \protect\\
Test-Time Adaptation for Unsupervised Time-Series Anomaly Detection}

\author {
    Dongmin Kim, %\textsuperscript,{\rm 1},
    Sunghyun Park, %\textsuperscript,{\rm 1},
    Jaegul Choo %\textsuperscript,{\rm 1}
}

\affiliations {
    KAIST\\
    tommy.dm.kim@kaist.ac.kr, 
    psh01087@kaist.ac.kr, 
    jchoo@kaist.ac.kr
}

\usepackage{bibentry}
\begin{document}

\maketitle

\begin{abstract}
    Time-series anomaly detection deals with the problem of detecting anomalous timesteps by learning normality from the sequence of observations. 
    However, the concept of normality evolves over time, leading to a "new normal problem", where the distribution of normality can be changed due to the distribution shifts between training and test data. 
    This paper highlights the prevalence of the new normal problem in unsupervised time-series anomaly detection studies. 
    To tackle this issue, we propose a simple yet effective test-time adaptation strategy based on trend estimation and a self-supervised approach to learning new normalities during inference. 
    Extensive experiments on real-world benchmarks demonstrate that incorporating the proposed strategy into the anomaly detector consistently improves the model's performance compared to the baselines, leading to robustness to the distribution shifts.
\end{abstract}
\section{Introduction}
In real-world monitoring systems, the continuous operation of numerous sensors generates substantial real-time measurements. 
Time-series anomaly detection aims to identify observations that deviate from the concept of normality~\cite{DeepShallow, DLADR} within a sequence of observations. 
Examples of anomalous events include physical attacks on industrial systems~\cite{SWaT, MADGAN}, unpredictable robot behavior~\cite{LSTMVAE}, faulty sensors from wide-sensor networks~\cite{WangKD15, RassamMZ18}, 
cybersecurity attacks on server monitoring systems~\cite{OmniAnomaly,PSM}, and spacecraft malfunctions based on telemetry sensor data~\cite{HundmanCLCS18, ITAD, liu2016detection}.

However, detecting abnormal timesteps presents significant challenges due to several factors. 
Firstly, the complex nature of system dynamics, characterized by the coordination of multiple sensors, complicates the task.
Secondly, the increasing volume of incoming signals to monitoring systems adds to the difficulty. 
Lastly, acquiring labels for abnormal behaviors is problematic. 
To address these challenges, unsupervised time-series anomaly detection models~\cite{AnomalyTransformer, USAD, OmniAnomaly, LSTMVAE, LSTMEncDec} have emerged, focusing on learning normal patterns from available training datasets and being deployed after training.
 
\begin{figure*}[t!]
    \centering
    \includegraphics[width=1.0\linewidth]{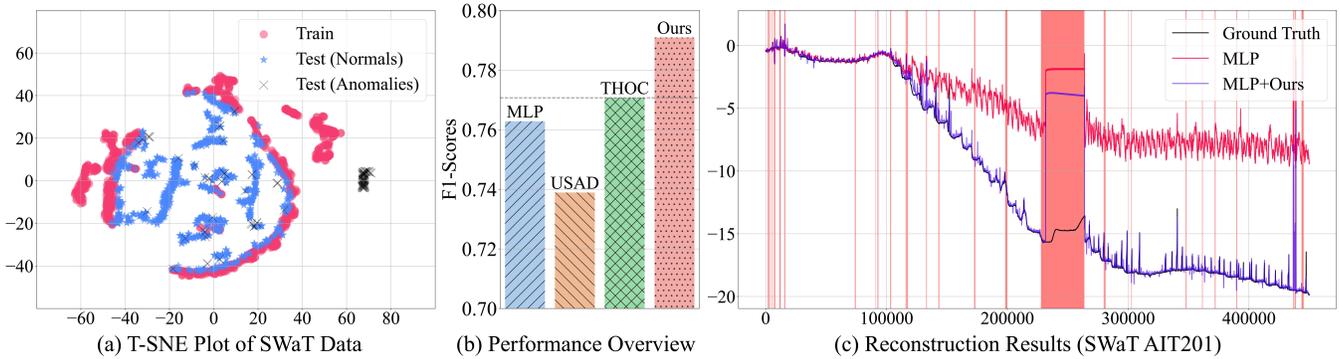}
    \caption{
    Motivation for learning new normals. (a) T-SNE visualization of the SWaT benchmark~\cite{SWaT} reveals distinct behavior between the training (red) and test data (blue). 
    (b) Our test-time adaptation strategy surpasses previous state-of-the-art time-series anomaly detection models in terms of F1 score, even with simple baselines such as MLP-based autoencoders. (c) This improvement arises from effectively handling significant distribution shifts in the time-series data. Over time, off-the-shelf models fail to adapt to these new normals, while our approach exhibits robustness to such distribution shifts. Consequently, previous approaches~\cite{USAD, THOC} produce false positive cases due to the model's inability to keep pace with changing dynamics, thereby \emph{"the model is staying in the past while the world is changing."}
    }
\label{fig:Main_Figure}
\end{figure*}

Nevertheless, the concept of normality can change over time, widely known as a distribution shift~\cite{DatasetShift, RevIN, TTT, DomainBed, TENT, CoTTA}, as can be seen in the Fig.~\ref{fig:Main_Figure}-(a). 
We have observed that off-the-shelf models are susceptible to such shifts, leading to a "new normal problem", where the distribution of normality during test time cannot be fully characterized solely based on training data. 
Without consideration of distribution shifts, these models tend to rely on past observations and generate false alarms, compromising the consistency of monitoring systems~\cite{Anoshift, ADunderDS}.

% Figure - Main Motivation

Recently, test-time adaptation mechanisms~\cite{TENT, CoTTA, niu2022efficient} have been proposed to adapt models for alleviating performance degradation due to distribution shifts between training and test datasets, especially in the computer vision field. 
Test-time adaptation methods update the model parameters to generalize to different data distributions, without relying on either additional supervision from labels or access to training data. 
Time-series anomaly detection task also shares motivation for applying test-time adaptation strategies; frequent access to past data for adaptation is costly as monitoring systems work in real-time~\cite{PSM, ITAD, OmniAnomaly} and model update without supervision is desired as acquiring labels is often limited~\cite{TadGAN, DeepShallow, USAD}. 
Motivated by these advancements, we propose a test-time adaptation for unsupervised time-series anomaly detection under distribution shifts.

Our paper highlights the prevalence of the new normal problem in time-series anomaly detection literature. 
To address this issue properly, we propose a simple yet effective adaptation strategy using trend estimates and model updates with normal instances based on the model's prediction itself.
Trend estimate, given by the exponential moving average of the observations, follows the expected value of a time-series with adaptation to changing conditions~\cite{muth1960optimal} with computational efficiency. 
After model deployment, we update the model parameters with the normalized input sequence, which is detrended by subtracting the trend estimate, to learn complicated dynamics that cannot be captured solely on the trend estimate.
Our proposed method makes the model robust to such distribution shifts, thereby increasing detector performance, as shown in Fig.~\ref{fig:Main_Figure}-(b) and Fig.~\ref{fig:Main_Figure}-(c).

Our contributions can be summarized as follows:
\begin{itemize}[noitemsep, leftmargin=0.25in,topsep=0pt]
    \item We discover that new normal problems pose a significant challenge in modeling unsupervised time-series anomaly detection under distribution shifts.
    \item We propose a simple yet effective adaptation strategy following the trend estimate of the time-series data and update the model parameters using a detrended sequence to address these problems.
    \item Through extensive experiments on various real-world datasets, our method consistently improves the model's performance when facing a severe distribution shift problem between training data and test data.
\end{itemize}

\section{Related Works}

\noindent\textbf{Unsupervised time-series anomaly detection.}
Unsupervised time-series anomaly detection~\cite{OmniAnomaly, USAD, AnomalyTransformer} aims to detect observations that deviate considerably from normality, assuming the non-existence of the available labels. 
To the extent of conventional anomaly detection approaches~\cite{LOF, OCSVM, SVDD} and deep-learning-based anomaly detection approaches~\cite{DAGMM, DeepSVDD}, unsupervised time-series anomaly detection models aim to build an architecture that can model the temporal dynamics of the sequence. 

The main categories of unsupervised time-series anomaly detection models include reconstruction-based models, clustering-based models, and forecasting-based models.
Building upon the assumption of better reconstruction performance of normal instances compared to anomalous instances, reconstruction-based models encompass a range of approaches involving LSTM~\cite{LSTMEncDec, LSTMVAE, OmniAnomaly} and MLP~\cite{USAD} architectures, as well as the integration of GANs~\cite{AnoGAN, TadGAN, MADGAN}.
Clustering-based methods include the extension of one-class support vector machine approaches~\cite{OCSVM, SVDD}, tensor decomposition-based clustering methods for the detection of anomalies~\cite{ITAD}, and the utilization of latent representations for clustering~\cite{DeepSVDD, THOC}.
Forecasting-based methods rely on detecting anomalies by identifying substantial deviations between past sequences and ground truth labels, as exemplified by the use of ARIMA~\cite{PenaAP13}, LSTM~\cite{HundmanCLCS18}, and transformer~\cite{AnomalyTransformer}.

% Figure - Methods illustration
\begin{figure}[t!]
    \centering
    \includegraphics[width=1.0\linewidth]{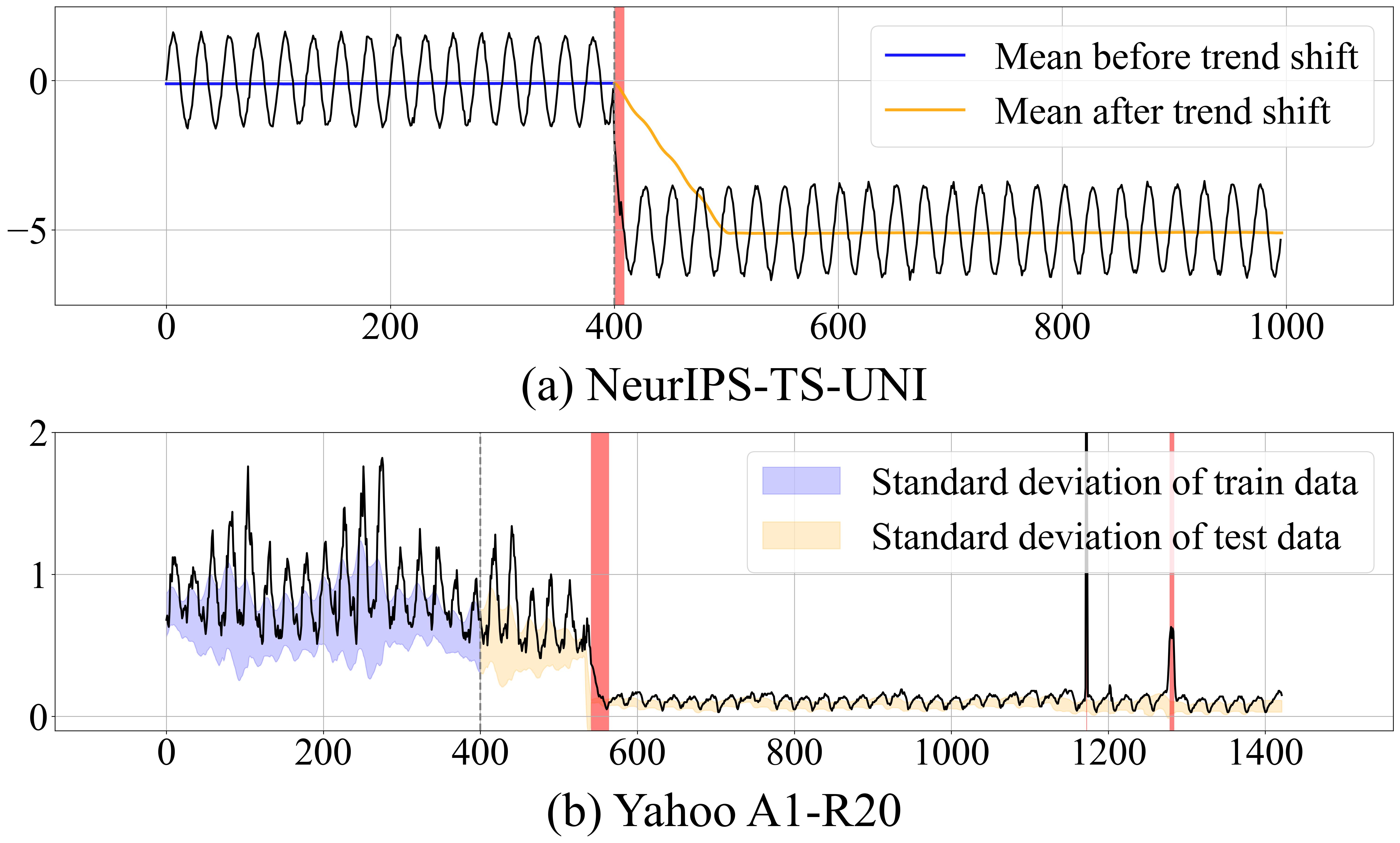}
    \caption{
    Illustration of the necessity for estimating trends and test-time adaptation. (a) NeurIPS-TS-UNI shows synthetic data generated based on the previous work~\cite{NTS}, revealing an abrupt trend shift while preserving underlying dynamics. The objective of the trend estimation module is to adapt to such trend shifts successfully. 
    (b) Solely relying on trend estimation may not be adequate to fully capture the dynamics, as demonstrated by the Yahoo-A1-R20 series. The shaded purple and yellow areas represent the standard deviations of the train and test data, respectively. To model this shift in dynamics, which cannot be fully captured alone with trend estimates, it is necessary to learn distribution shifts through test-time model updates outlined directly.
    }
\label{fig:main_ion}
\end{figure}

% distribution shift
\noindent\textbf{Distribution shift in time-series data.}
Due to the nature of continually changing temporal dynamics, mitigating distribution shifts emerges as a pivotal concern within the time-series data analysis, notably within tasks such as time-series forecasting~\cite{RevIN, NST} and anomaly detection~\cite{FITNESS, Anoshift}.

Online RNN-AD \cite{Online-RNN-AD} adapts to concept drift with RNN architectures, which update the model with backpropagation of anomaly scores using all stream data. 
Our work differentiates from this work by introducing detrending modules for model updates and selective learning of a set of normal instances in a self-supervised way.
Although recent work~\cite{FITNESS} also presents an adaptable framework for anomaly detection, it hinges on a dynamic window mechanism applied to historical data streams.
Notably, our approach diverges from their assumption of accessibility of past sequences; we keep model parameters at hand, process input sequences immediately, and evict after.

% TTA
\noindent\textbf{Test-time adaptation.}
To alleviate the performance degradation caused by distribution shift, unsupervised domain adaptation~\cite{ganin2016domain, zou2018unsupervised, yoo2022unsupervised, liang2020we} methods have been developed in various fields. 
These methods align with our work from the perspective of addressing the covariate shift problem.
In recent times, fully test-time adaptation (TTA)~\cite{TENT} methods have emerged to enhance the model performance on test data through real-time adaptation using unlabeled test samples during inference, without relying on access to the training data.
Most TTA approaches employ entropy minimization~\cite{TENT, niu2022efficient, choi2022improving} or pseudo labels~\cite{CoTTA} to update the model parameters using unlabeled test samples.
However, simply adopting previous TTA methods may not be directly applicable to unsupervised time-series anomaly detection.
This is due to the vulnerability of the model when updating the model using all test samples, as abnormal test samples have the potential to disrupt its functionality.
Consequently, this work aims to successfully apply the concept of test-time adaptation to the unsupervised time-series anomaly detection task.

\section{Method}
\label{sec:method}

\subsection{Problem Statement}
\label{sec:method:problem_statement}
% Anomaly detection overall description
Unsupervised time-series anomaly detection aims to detect anomalous timesteps during test time without explicit supervision by learning the concept of normality.
The concept of normality is defined as the probability distribution $\mathbb{P}$ on data $\mathcal{D}$ that is the ground-truth law of normal behavior in a given task~\cite{DeepShallow}.
Accordingly, a set of anomalies is defined as data with sufficiently small probability under such distribution, \ie, $p(x) < \epsilon $.
New normal problem that we tackle can be formulated as the phenomena of underlying distribution $\mathbb{P}$ is not stationary, \ie, $\mathbb{P}_{train} \neq \mathbb{P}_{test}$.

% time series data
For observations over $N$ timesteps with $F$ features, time-series data is specified by a sequence $\mathcal{D} = \{X_1, X_2, ..., X_N\}$, where each $X_i \in \mathbb{R}^F$.
An anomaly detector aims to map each observation to a class label $y=\{0, 1\}$, where $y=0$ and $y=1$ each denote normal and abnormal timesteps. 
The detector is specified by an anomaly score function $\mathcal{A}: \mathbb{R}^F \rightarrow \mathbb{R}$, along with a decision threshold $\tau$.
Concretely, observation $X_t$ is classified as anomalous if $\mathcal{A}(X_t) > \tau$.
We denote the set of train-time instances as $\mathcal{D}_{train}$ and the set of test-time instances as $\mathcal{D}_{test}$.
Accordingly, test-time normals and anomalies can be defined each as $\{X \in \mathcal{D}_{test}\ |\ y=0\}$ and $\{X \in \mathcal{D}_{test}\ |\ y=1\}$.

% sliding window
To reflect the temporal context of time-series data to detect anomalous timestep(s), a set of observations $\mathcal{D}$ is preprocessed with a sliding window setting. 
Specifically, we denote a sequence of $w$ observations until timestep $t$ as $\mathcal{X}_{w, t} = [X_{t-w+1}, X_{t-w+2}, ..., X_{t-1}, X_{t}]$ and its corresponding class label and prediction of the model as $\mathcal{Y}_{w,t} = [y_{t-w+1}, y_{t-w+2}, ..., y_{t-1}, y_{t}]$ and $\hat{\mathcal{Y}}_{w,t} = [\hat{y}_{t-w+1}, \hat{y}_{t-w+2}, ..., \hat{y}_{t-1}, \hat{y}_{t}]$ following conventional approaches of the time-series anomaly detection literatures~\cite{THOC, OmniAnomaly}.

\subsection{Input Normalization Using Trend Estimate}
\label{sec:method:Trend_New_Normals}

\begin{figure}
    \centering
    \includegraphics[width=0.8\linewidth]{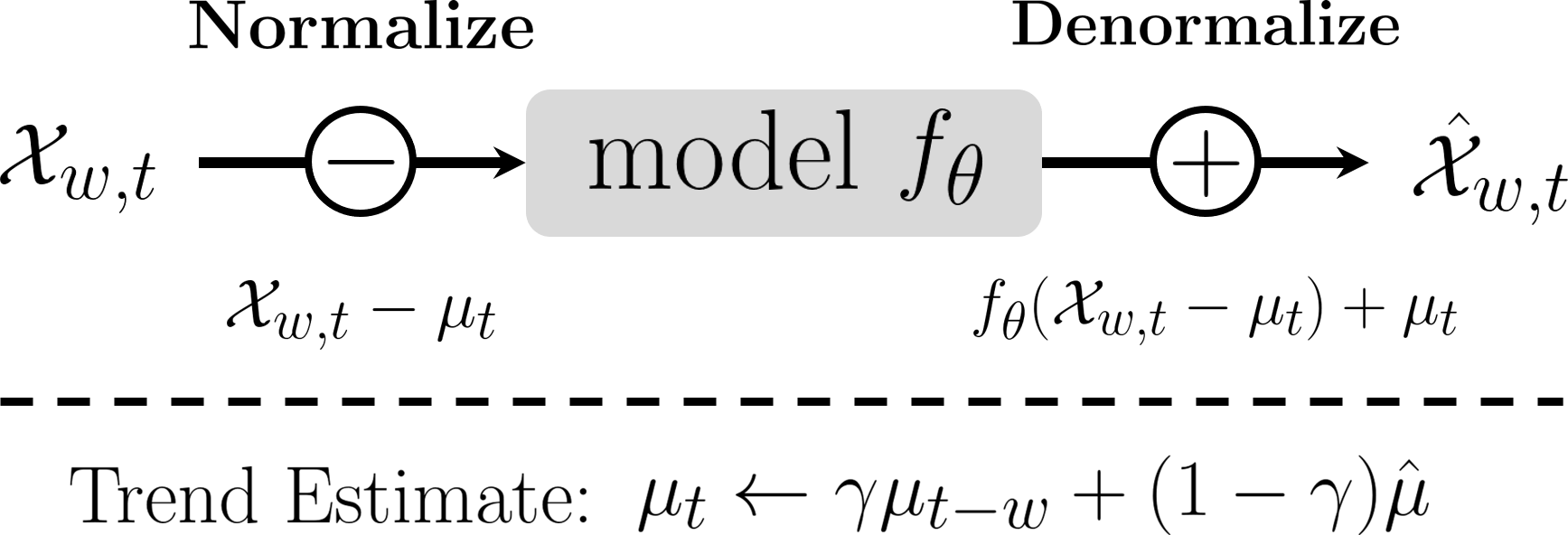}
    \caption{Illustration on detrend module.}
    \label{fig:detrend_module}
\end{figure}

A trend estimation module aims to adapt to new normals that significantly differ in trend with preserving the underlying dynamics of the sequence. 
Accordingly, previous work~\cite{NTS} defines trend-outlier as:
\begin{equation}
    \Delta (\mathcal{T}(\cdot), \tilde{\mathcal{T}}(\cdot)) > \delta,    
\end{equation}
where $\Delta$ is a function that measures the discrepancy between two functions. $\tilde{\mathcal{T}}$ is a function that returns the trend of normal sequences. $\mathcal{T}$ is a trend of an arbitrary sequence to compare to the trend of normal sequences. Fig.~\ref{fig:main_ion}-(a) illustrates the importance of properly estimating the trend of normalities. Even though sequences before and after the transition shares the same dynamics, observations after the trend shift are classified as anomalies without proper adaptation to trends.
To address such a problem, we simply detrend with trend estimates using the exponential moving average statistics. Technically, we estimate the trend as:
\begin{equation}
    \tilde{\mathcal{T}}(\cdot) : \mu_{t} \leftarrow \gamma\mu_{t-w} + (1-\gamma)\hat{\mu},    
\end{equation}
where $\hat{\mu} = \frac{1}{w}\Sigma_{i=t-w+1}^{t} X_i$, which is the empirical mean of the stream data, and $\gamma$ is a hyperparameter that controls an exponentially moving average update rate for tracking the trend of the data stream. This procedure is one form of eliminating nonstationary trend components with mean adjustment~\cite{Shumway_Stoffer_2017}, allowing models to be updated with numerical stability. 
Concretely, as shown in Fig.~\ref{fig:detrend_module}, along with reconstruction-based anomaly detection models, the model reconstructs detrended sequence $\mathcal{X}_{w, t} - \mu_t$ instead of $\mathcal{X}_{w, t}$ and denormalize the reconstructed sequence by adding estimated trend for the final output.

\subsection{Model Update with New Normals}
\label{sec:method:NNLearning}

Test-time adaptation with model update aims to learn the underlying dynamics of the time series data, which cannot be fully captured by trend estimation alone, as shown in Fig.~\ref{fig:main_ion}-(b). 
Specifically, our approach continuously updates the model parameters with normal sequences during test time in a fully unsupervised manner. 
Formally, the normal instances during test-time observations can be formulated as $\{X \in \mathcal{D}_{test}\ |\ y=0\}$.
%$X_t \in \mathcal{X}_{test} \cap \mathcal{X}_{\nu}$.
To update the model parameters $\theta$ during test-time, the prediction of the model itself, $\hat{\mathcal{Y}}$ acts as selection criteria for filtering normal timesteps. The model is updated based on online gradient descent~\cite{OGD} using the following scheme:
\begin{equation}
    \theta \leftarrow \theta - \eta\nabla_{\theta}\mathcal{L}(\mathcal{X}_{w,t}, \hat{\mathcal{Y}}_{w, t}, \mu_{t}, \tau),    
\end{equation}
where $\eta$ is the test-time learning rate for the model update. 
$\tau$ denotes a threshold for classifying the anomalous timesteps. 
Specifically, our approach uses autoencoder architectures along with reconstruction loss. Hence, mentioned updating scheme can be further described as:
% \begin{equation}
%     \mathcal{L}(\mathcal{X}_{w,t}, \hat{\mathcal{Y}}_{w,t}, \mu_{t}, \tau) = {{\sim}\hat{\mathcal{Y}}_{w,t}^T}(\hat{\mathcal{X}}_{w,t} - \mathcal{X}_{w,t})^2,   
% \end{equation}
\begin{equation}
    \mathcal{L}(\mathcal{X}_{w,t}, \hat{\mathcal{Y}}_{w,t}, \mu_{t}, \tau) = (1 - \hat{\mathcal{Y}}_{w,t}^\top)(\hat{\mathcal{X}}_{w,t} - \mathcal{X}_{w,t})^2,   
\end{equation}
where $\hat{\mathcal{X}}_{w,t}$ denotes reconstructed output from the model and ${\hat{\mathcal{Y}}}_{w,t}$ denotes predicted labels, where 0 and 1 indicate normal and abnormal, respectively.
% where $\hat{\mathcal{X}}_{w, t}$ denotes reconstructed output from the model and ${{\sim}\hat{\mathcal{Y}}}$ denotes flip operation that maps each prediction from 0 to 1 and vice versa.

Although we utilize the entire time-series data for trend estimate, we only incorporate the normal instances to update the model based on the model's predictions.
The rationale behind this strategy stems from the assumption that unsupervised anomaly detectors are trained using normal data before model deployment.
Consequently, the inclusion of anomaly samples for model updates during test time can potentially have a detrimental impact on the model's performance.
In contrast, to enable trend estimation even in scenarios with substantial variations, it is essential to incorporate normal instances that could potentially be predicted as anomalies by the anomaly detector.
\section{Experiments}
\label{sec:experiments}

%%%%%%%%%%%% Experiment Setups %%%%%%%%%%%%
\subsection{Experiment Setups}

% Dataset
% 간결하게 하고, appendix로.

\noindent\textbf{Datasets.} 
We selected datasets for experiments based on the following criteria: (i) widely used datasets in time-series anomaly detection literature (SWaT), (ii) subsets with significant distribution shifts from commonly utilized datasets (SMD, MSL, SMAP), (iii) datasets including substantial distribution shifts (WADI, Yahoo), (iv) datasets with minimal distribution shifts (CreditCard).

Descriptions for the real-world datasets we utilized are as follows.
% 우리는 다음과 같은 기준으로 실험에 사용할 데이터셋을 선정하였다.: 기존의 anomaly detection literature에서 많이 사용하는 데이터셋 (SwaT)와 많이 사용되는 데이터셋에서 distribution shift가 큰 subset (SMD, SMAP)과 distribution shift가 큰 데이터셋 (WADI, Yahoo)와 distribution shift가 적은 데이터셋 (CreditCard)를 선정하여 실험을 진행하였다.  
% 우리가 기존의 anomaly detection literature에서 많이 사용되는 dataset들 (SWaT, SMD, SMAP)과 distribution shift가 큰 데이터셋과 (WADI, Yahoo) distribution shift가 적은 데이터셋 (CreditCard) 로 실험을 했다. 이런말을 넣으면 좋을듯!
(1) The SWaT~\cite{SWaT} and WADI~\footnote{iTrust, Centre for Research in Cyber Security, Singapore University of Technology and Design} consist of measurements collected from water treatment system testbeds. SWaT dataset covers 11 days of measurement from 51 sensors, while WADI dataset covers 16 days of measurement from 123 sensors.
(2) The SMD dataset~\cite{OmniAnomaly} includes 5 weeks of data from 28 distinct server machines with 38-dimensional sensor inputs. For the experiment, two specific server machines (Machine 1-4 and Machine 2-1) were selected due to their distribution shift problems.
(3) The SMAP and MSL~\cite{HundmanCLCS18} datasets are derived from spacecraft monitoring systems. SMAP dataset comprises monitoring data from 28 unique machines with 55 telemetry channels, whereas MSL dataset includes data from 19 unique machines with 27 telemetry channels. Data from two specific machines with distribution shifts, MSL (P-15) and SMAP (T-3), are selected for our experiments.
(4) The CreditCard dataset~\footnote{https://www.kaggle.com/datasets/mlg-ulb/creditcardfraud} consists of transactional logs spanning two days. It contains 28 PCA-anonymized features along with time and transaction amount information.
(5) The Yahoo dataset~\footnote{https://webscope.sandbox.yahoo.com/} is a combination of real (A1) and synthetic (A2, A3, A4) datasets. Yahoo-A1 dataset contains 67 univariate real-world datasets, with a specific focus on two datasets (A1-R20 and A1-R55) exhibiting distribution shift problems. 
Further details and main statistics of the datasets can be found in the supplementary. % Appendix~\ref{appendix:DD}.

% Figure - KL Divergence of Datasets
\begin{figure}[t!]
    \centering
    \includegraphics[width=1\linewidth]{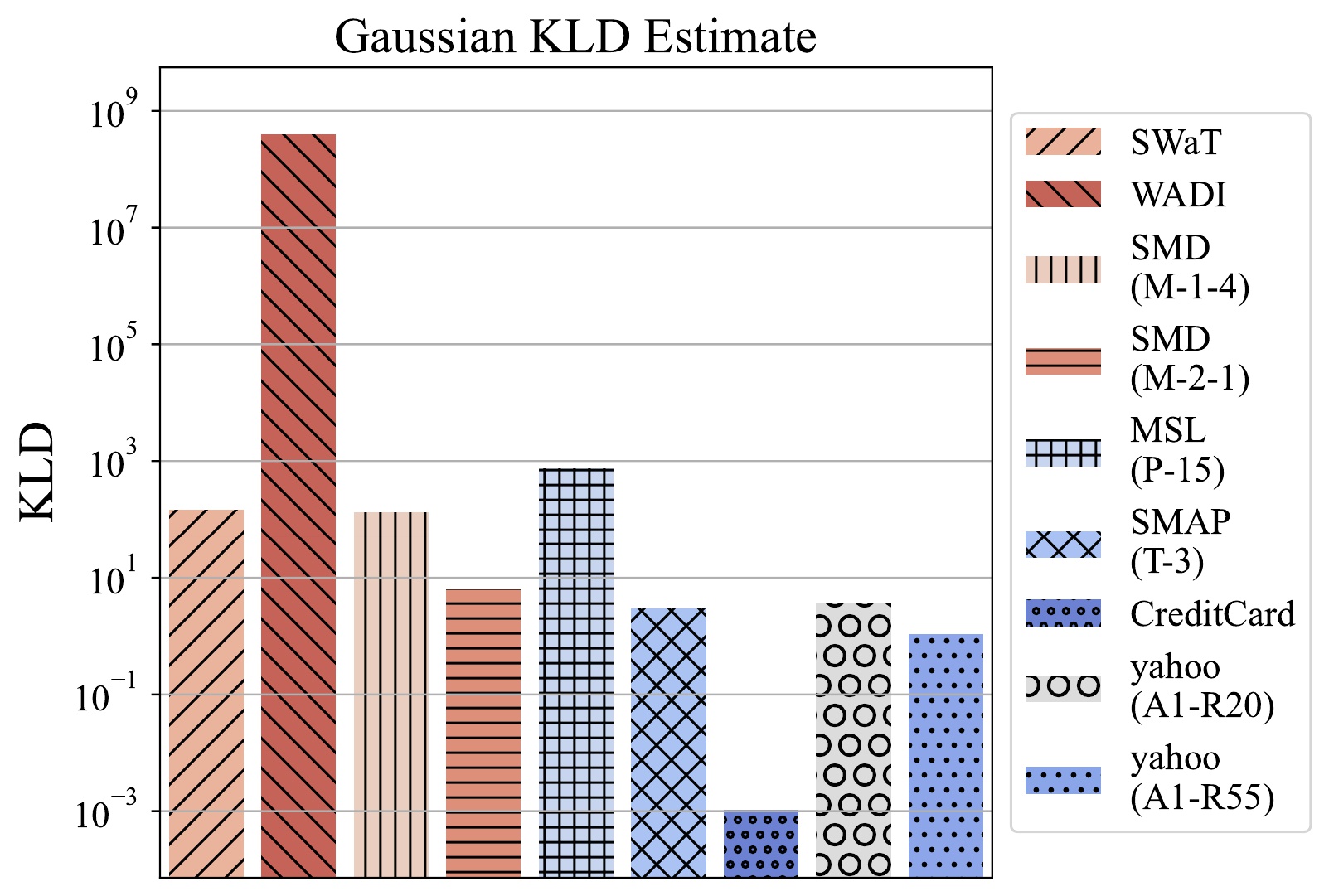} 
    \caption{
    Kullback–Leibler Divergence (KLD) of various datasets. 
    $D_{KL}(\mathcal{D}_{test} || \mathcal{D}_{train})$ is given, which implies how much additional information is needed to fully describe $\mathcal{D}_{test}$, given $\mathcal{D}_{train}$. The measure quantifies the distribution shift problem of the datasets.}
    \label{fig:main_kld}
\end{figure}

% Baselines: 
% official implementation default 를 썼다
% 어떤 세팅으로 학습을 시켰는지?? (Unsupervised / Normal로 %가정하고) / Test는 어떻게 했다
% 중요한 점: 우리가 왜 이런 세팅을 채택했는가에 대한 근거 / 논리 잘 설명되어야함!
\noindent\textbf{Baselines.} 
We compare our methodology with 5 baselines: MLP-based autoencoder (MLP), LSTMEncDec (LSTM)~\cite{LSTMEncDec}, USAD~\cite{USAD}, THOC~\cite{THOC} and anomaly transformer (AT)~\cite{AnomalyTransformer}. 
LSTM, USAD, and THOC have been re-implemented based on the description of each paper. 
Official implementation of anomaly transformer\footnote{https://github.com/thuml/Anomaly-Transformer} is utilized in our experiments. 
We use hyperparameters and default settings of THOC, USAD, and AT described in their papers. 
MLP and LSTM use the latent dimension of 128 as default. 
As all the approaches are fully unsupervised, we trained all the models with the assumption of normality for train datasets. 
During test time, our approach gets input of $w$ non-overlapping window, which is the same input as the train-time window size. 
Details of hyperparameters can be found in supplementary. %Appendix~\ref{appendix:HypSet}.

% Evaluation Metrics
% AUROC 쓴 이유를 조금 더구체적으로
\noindent\textbf{Evaluation metrics.} 
We report a metric called F1-PA~\cite{XuCZLBLLZPFCWQ18}, widely utilized in the recent time-series anomaly detection studies~\cite{AnomalyTransformer, THOC, USAD, OmniAnomaly}. 
This metric views the whole successive abnormal segment as correctly detected if any of the timesteps in the segment is classified as an anomaly. 
Note that F1-PA metric overestimate classifier performance~\cite{KimCCLY22}, even though this metric has practical justifications~\cite{XuCZLBLLZPFCWQ18}.

Therefore, we consider three additional evaluation metrics, which are F1 score, area under receiver operating characteristic curve (AUROC), and area under the precision-recall curve (AUPRC).
Different from F1-PA, the F1 score can measure the anomaly detection status for each individual timestep, which directly reflects the performance of the anomaly detector.
We also report AUROC and AUPRC over test data anomaly scores, which gives an overall summary of anomaly detector performance for all possible candidates of thresholds $\tau$.
AUROC takes into account the performance across all possible decision thresholds, making it less sensitive to the choice of a specific threshold. 
We measure AUPRC, which is well-suited for imbalanced classification scenarios~\cite{saito2015precision, NavMetricMaze}.

% There also has been a proposition that AUROC can be misleading, especially for imbalanced classification scenarios~\cite{saito2015precision, NavMetricMaze}, AUPRC is measured as an alternative metric.
% As mentioned in \ref{sec:method:problem_statement}, anomaly datasets, including datasets of this paper, have highly imbalanced properties, with the importance of correctly detecting anomalous samples. Therefore, the F1-score, which is defined as a harmonic mean of precision and recall scores, is reported to evaluate the classifier performance correctly.

% We also report metrics with adjustment, called F1-PA, which is firstly suggested by Xu~\etal~\cite{XuCZLBLLZPFCWQ18} and widely used in the recent time-series anomaly detection studies~\cite{AnomalyTransformer, THOC, USAD, OmniAnomaly}.
% Moreover, 

% In addition to F1 and F1-PA, we report the area under receiver operating characteristic curve (AUROC) over test data anomaly scores, which gives an overall summary of anomaly detector performance for all possible candidates of thresholds $\tau$.

For brevity, we report these four metrics in the main paper. 
Other metrics for adjusted and non-adjusted metrics, including accuracy, precision, recall, F1, and confusion matrix (The number of true negatives, false positives, false negatives, and true positives), are provided in the supplementary.

%%%%%%%%%%%% MAIN %%%%%%%%%%%%
% consistent 하게, 전반적으로 심플한 방법임에도 성능을 잘 올린다
%- 성능 떨어진걸 잘 커버
% (역전된 경우) ex wadi f1-pa 잘 서술하기
% (1) Comparison with Baselines.
% (2) AUROC Curve Visualization
\subsection{Comparison with Baselines}
\label{main_exp}
\textbf{Main results.} 
To validate the effectiveness of our method, we conducted a comparative analysis between unsupervised time-series anomaly detection models and the MLP model combined with our approach. 
As presented in Table~\ref{table:Main_Exp}, the results demonstrate that our method consistently improves the performance of the MLP model across various evaluation metrics. 
Notably, we achieve a significant improvement of up to 13\% in the AUROC of the WADI dataset and 51\% in the AUPRC of MSL (P-15), which exhibits a distribution shift problem as illustrated in Fig.~\ref{fig:main_kld}.
In the case of the Yahoo A1-R20 dataset, shown in Fig.~\ref{fig:main_ion}-(b), our method demonstrates the highest performance gain in terms of the F1 score.
In contrast to most of the datasets, our method shows only marginal improvement in the CreditCard dataset.

It is due to the fact that the dataset has a minimal distribution shift problem, resulting in a limited performance gain. 
The dataset that exhibits lower F1 performance compared to the off-the-shelf baseline is WADI. 
This discrepancy is a result of the threshold setting with test anomaly scores. 
Specifically, the maximum anomaly score for the WADI train data using the USAD model is 0.225, while the threshold that yields the reported F1 score in the table is 585.845, which is significantly higher. 
Consequently, although USAD and LSTM models exhibit higher scores for F1, the overall classifier performance measured by AUROC is lower. 

Moreover, we compared our method to the anomaly transformer (AT), one of the state-of-the-art methods. 
While AT shows comparable performance in terms of F1-PA, it falls short regarding the F1 score, AUROC, and AUPRC. 
This disparity arises because the anomaly transformer generates positive predictions at certain intervals rather than specifying the exact moments of anomalous points. 
Details of test-time anomaly scores of baselines are reported in supplementary.

\begin{table}[t!]
\begin{center}

\setlength{\tabcolsep}{2pt}
\small
\begin{tabular}{@{}l|l|ccccc|c}
\toprule
\multicolumn{1}{c|}{\textbf{Dataset}} & \textbf{Metrics} & \textbf{MLP} & \textbf{LSTM} & \textbf{USAD} & \textbf{THOC} & \textbf{AT} & \textbf{Ours} \\
\midrule\midrule

\multirowcell{4}{SWaT}

&F1&0.765 &0.401 &0.557 &0.776 &0.218 &\textbf{0.784} \\%0.019
&F1-PA&0.831 &0.768 &0.655 &0.862 &\textbf{0.962} &0.903 \\%0.072
&AUROC&0.832 &0.697 &0.737 &0.838 &0.530 &\textbf{0.892} \\%0.06
&AUPRC&0.722 &0.248 &0.457 &0.744 &0.195 &\textbf{0.780} \\%0.058
\midrule

\multirowcell{4}{WADI} 
&F1&0.131 &0.245 &\textbf{0.260} &0.124 &0.109 &0.148 \\%0.017
&F1-PA&0.175 &0.279 &0.279 &0.153 &\textbf{0.915} &0.346 \\%0.171
&AUROC&0.485 &0.525 &0.530 &0.484 &0.501 &\textbf{0.624} \\%0.139
&AUPRC&0.052 &0.195 &\textbf{0.205} &0.144 &0.059 &0.081 \\%0.029
\midrule

\multirowcell{4}{SMD\\(M-1-4)}

&F1&0.273 &0.282 &0.159 &0.379 &0.059 &\textbf{0.463} \\%0.19
&F1-PA&0.544 &0.500 &0.296 &0.521 &0.799 &\textbf{0.874} \\%0.33
&AUROC&0.805 &0.818 &0.673 &\textbf{0.869} &0.479 &0.845 \\%0.04
&AUPRC&0.169 &0.151 &0.103 &0.223 &0.034 &\textbf{0.354} \\%0.185
\midrule

\multirowcell{4}{SMD\\(M-2-1)}
&F1&0.236 &0.283 &\textbf{0.308} &0.295 &0.094 &0.249 \\%0.013
&F1-PA&0.814 &0.910 &0.922 &0.705 &0.866 &\textbf{0.974} \\%0.16
&AUROC&0.674 &0.727 & 0.738 &0.668 &0.498 &\textbf{0.764} \\%0.09
&AUPRC&0.190 & 0.251 &0.246 &0.161 &0.052 &\textbf{0.280} \\%0.09
\midrule

\multirowcell{4}{MSL\\(P-15)} 
&F1&0.263 &0.056 &0.060 &0.018 &0.071 &\textbf{0.440} \\%0.177
&F1-PA&0.848 &0.351 &0.097 &0.027 &0.437 &\textbf{0.944} \\%0.096
&AUROC&0.645 &0.617 &0.661 &0.332 &0.568 &\textbf{0.801} \\%0.156
&AUPRC&0.061 &0.012 &0.016 &0.005 &0.023 &\textbf{0.575} \\%0.514
\midrule

\multirowcell{4}{SMAP\\(T-3)}
&F1&0.095 &0.091 &0.044 &0.154 &0.042 &\textbf{0.218} \\%0.123
&F1-PA&0.992 &\textbf{0.998} &0.940 &0.747 &0.772 &0.708 \\%-0.284
&AUROC&0.510 &0.515 &0.500 &0.591 &0.490 &\textbf{0.617} \\%0.107
&AUPRC&0.044 &0.050 &0.031 &0.049 &0.017 &\textbf{0.111} \\%0.067
\midrule

\multirowcell{4}{Credit\\Card} 
&F1&0.127 &0.220 &\textbf{0.323} &0.138 &0.039 &0.135 \\%0.008
&F1-PA&0.145 &0.234 &\textbf{0.323} &0.148 &0.056 &0.151 \\%0.006
&AUROC&0.943 &0.930 &0.887 &0.770 &0.548 &\textbf{0.943} \\%0
&AUPRC&0.055 &0.109 &\textbf{0.234} &0.041 &0.007 &0.063 \\%0.008
\midrule

\multirowcell{4}{Yahoo\\(A1-R20)} 
&F1&0.067 &0.065 &0.277 &0.106 &0.098 &\textbf{0.678} \\%0.611
&F1-PA&0.259 &0.426 &0.695 &0.106 &0.185 &\textbf{0.895} \\%0.636
&AUROC&0.367 &0.394 &0.668 &0.198 &0.525 &\textbf{0.971} \\%0.604
&AUPRC&0.056 &0.057 &0.161 &0.067 &0.048 &\textbf{0.637} \\%0.581
\midrule

\multirowcell{4}{Yahoo\\(A1-R55)} 
&F1&0.366 &0.446 &0.281 &0.059 &0.010 &\textbf{0.633} \\%0.267
&F1-PA&0.424 &0.446 &0.320 &0.059 &0.010 &\textbf{0.744} \\%0.32
&AUROC&0.916 &0.877 &0.867 &0.875 &0.478 &\textbf{0.958} \\%0.042
&AUPRC&0.303 &0.242 &0.177 &0.019 &0.002 &\textbf{0.624} \\%0.321
\bottomrule

\end{tabular}

\caption{
Comparison with the existing baselines. All results are based on five independent trials. 
This table reports the average of five trials for each metrics. Complete results with confidence intervals are reported in the supplementary. % Appendix~\ref{appendix:Main_Experiment_Full}.
}

\label{table:Main_Exp}

\end{center}
\end{table}
% \input{Tables/Main_Experiment_230813_2124}

% AUROC에서 우리 method가 SMD (M-2-1)을 제외하고는 가장 좋다. -> 이게 의미하는것은??
% 기존의 method들은 threshold에 sensitive함에도 불구하고, (AUROC가 낮다는 것은) entire test data를 써서 threshold를 잡아왔다.
% 그렇기 때문에 F1과 F1-PA에서는 종종 our method보다 좋은 결과를 보이곤 하지만, threshold에 민감해서 real-world scenario에서는 robust하지 않다는 치명적인 limitation이 있다.
% our method를 합치면, ~~~장점이 있다는 것을 확인할 수 있다.

\begin{figure}[t!]
    \centering
    \includegraphics[width=0.95\linewidth]{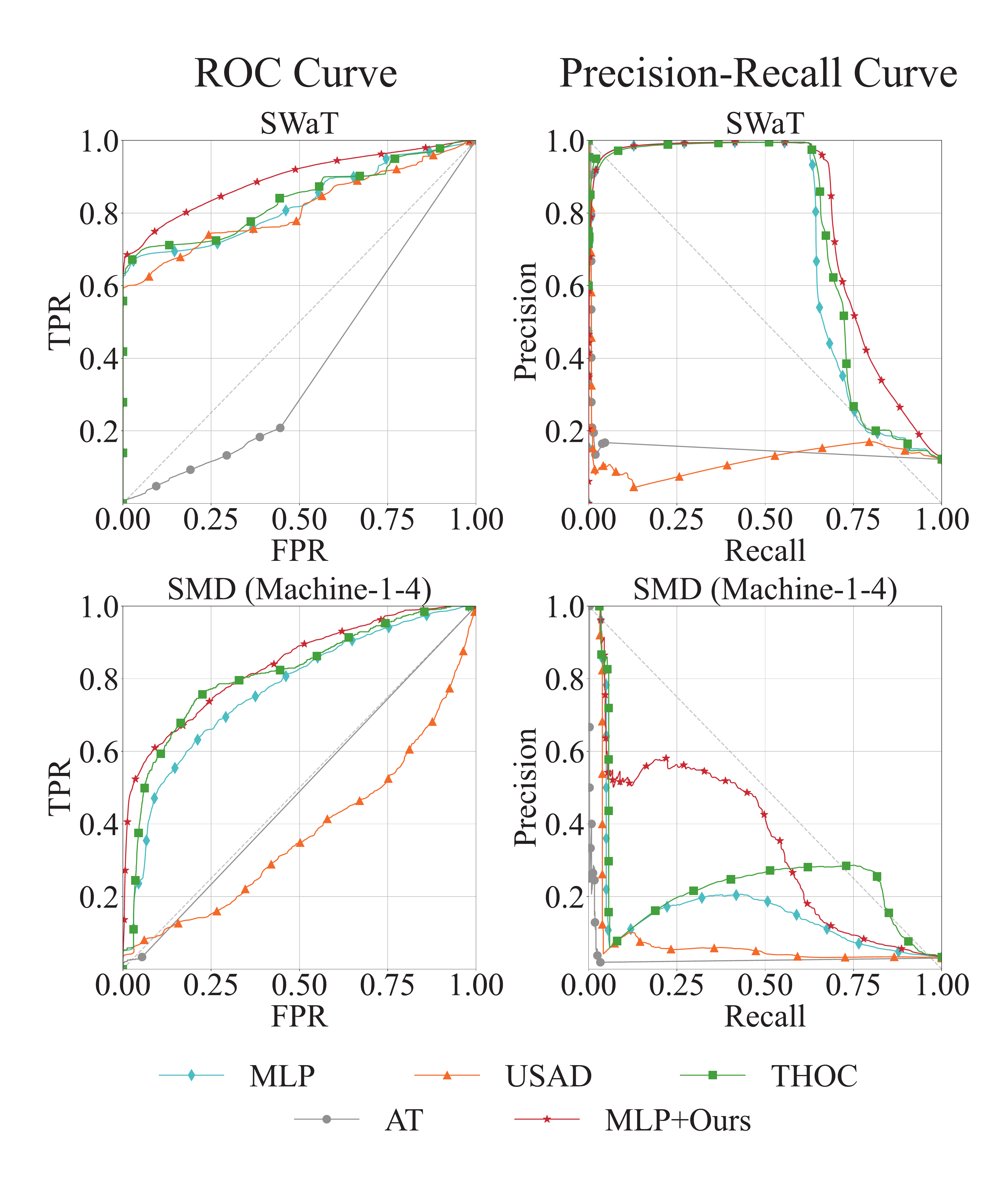}
    
    \caption{ROC curves (left) Precision-Recall curves (right) visualizations of baselines and MLP+Ours.}
\label{fig:main_roc}
\end{figure}

\noindent\textbf{Analysis on ROC and Precision-Recall curves.}
Our method consistently outperforms previous approaches in terms of AUROC across all datasets except for SMD (M-1-4), and AUPRC across all datasets except for WADI and Creditcard. 
This indicates that previous off-the-shelf baselines are sensitive to threshold settings, which poses a challenge for robustness in real-world scenarios where finding an optimal threshold is difficult.
Fig.~\ref{fig:main_roc} shows a visualization of the receiver operating curve (ROC curve) and precision-recall curve of our approach, along with baselines. Consistently, for both, our approach (red) improves the off-the-shelf classifier results (blue) significantly.

%%%%%%%%%%%% AnoShift %%%%%%%%%%%%
\subsection{Results on AnoShift Benchmark}
\label{casestudy}

\begin{figure}[t!]
    \centering
    \includegraphics[width=1.0\linewidth]{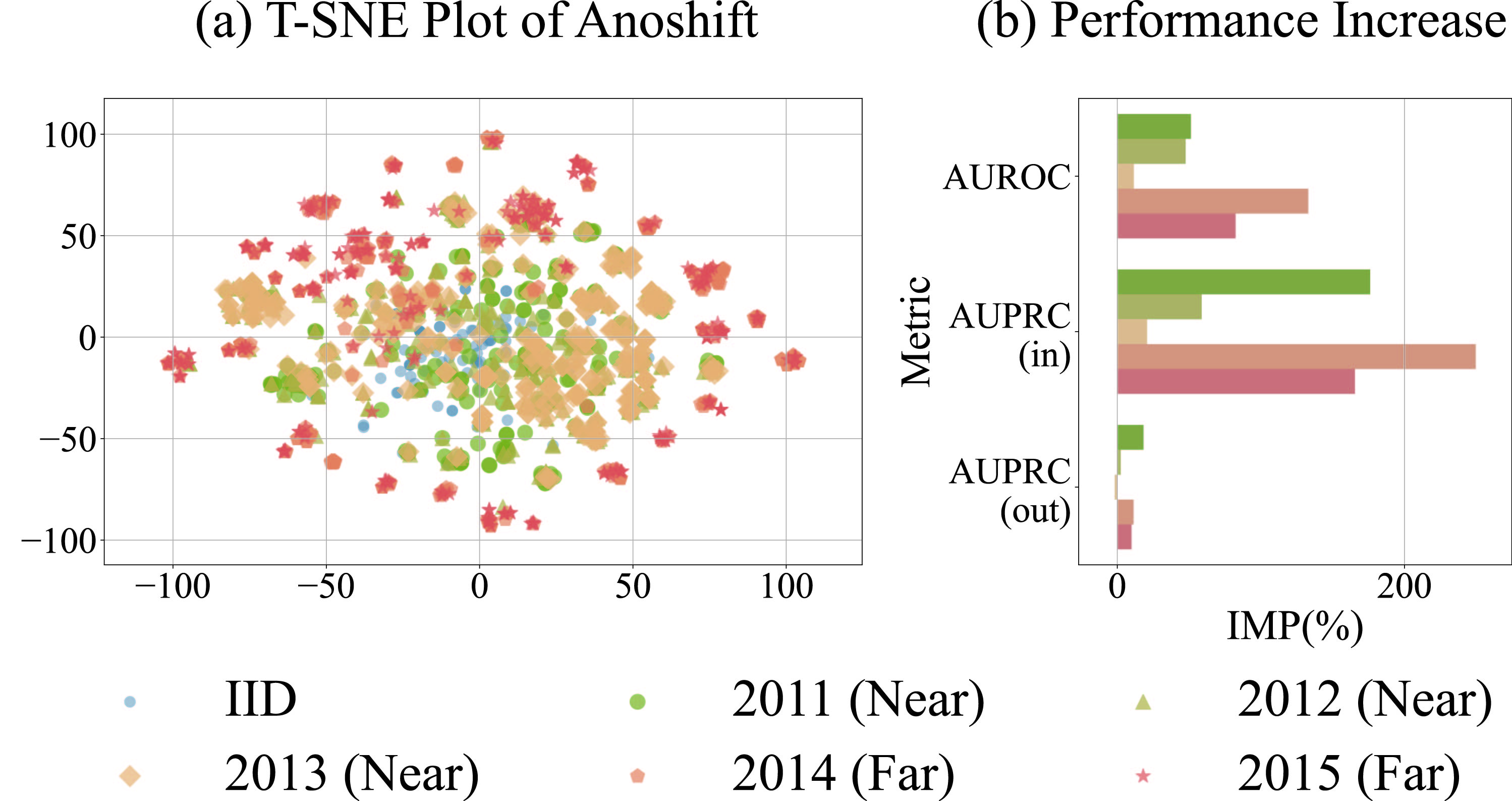}

    \caption{
    (a) T-SNE plot according to chronological distance and (b) performance increase with respect to three different evaluation metrics: AUROC, AUPRC-in and AUPRC-out.
    }
\label{fig:anoshift}
\end{figure}

The AnoShift benchmark~\cite{Anoshift} offers a testbed for the robustness of anomaly detection algorithm under distribution shift problem. The dataset spans a decade, partitioned into a training set covering the period 2006-2010, and two distinct test sets denoted as NEAR (2011-2013) and FAR (2014-2015). Visualized in Fig.~\ref{fig:anoshift}-(a), the data distribution progressively deviates from the train set as time progresses.

The principal objective of evaluation on the AnoShift benchmark is to investigate the effectiveness of our proposed algorithm against such distribution shifts. 
The evaluation entails three metrics—namely, Area Under the Receiver Operating Characteristic curve (AUROC), Area Under the Precision-Recall Curve with inliers as the positive class (AUPRC-in), and Area Under the Precision-Recall Curve with outliers as the positive class (AUPRC-out), following previous work~\cite{Anoshift}.
The performance of our method is compared to other deep-learning-based baselines, including SO-GAAL~\cite{SO-GAAL}, deepSVDD~\cite{DeepSVDD}, LUNAR~\cite{LUNAR}, ICL~\cite{ICL}, BERT~\cite{BERT} for anomalies.

Table~\ref{table:anoshift} demonstrates a significant improvement in performance when our method is integrated into an MLP-based autoencoder, as evidenced by an increase in AUROC of up to 0.216. Despite its simplicity, our approach markedly augments the baseline MLP performance, which previously showed inferior performance. This improvement is especially significant in FAR splits, which entail a severe distribution shift problem compared to NEAR splits. While our experiments focused on MLP, it's worth noting that our module can be seamlessly added to other baselines.
% 성능이 다른 모델이 SOTA인거에 대해서는 설명할 필요가 있을것 같긴한데

\begin{table}[t!]
\begin{center}

\setlength{\tabcolsep}{2pt}
\footnotesize
% \scriptsize
\begin{tabular}{@{}l|ccc|ccc@{}}
\toprule
\multirow{3}{*}{\textbf{Method}} 
& \multicolumn{3}{c|}{\textbf{NEAR}}
& \multicolumn{3}{c}{\textbf{FAR}}\\
\cmidrule{2-4}\cmidrule{5-7}
% & \multirowcell{2}{AUROC} & \multirowcell{2}{AUPRC\\(in)} & \multirowcell{2}{AUPRC\\(out)}
% & \multirowcell{2}{AUROC} & \multirowcell{2}{AUPRC\\(in)} & \multirowcell{2}{AUPRC\\(out)}  
& \multirowcell{2}{ROC} & \multirowcell{2}{PRC\\(in)} & \multirowcell{2}{PRC\\(out)}
& \multirowcell{2}{ROC} & \multirowcell{2}{PRC\\(in)} & \multirowcell{2}{PRC\\(out)}  
\\ 
&&&& \\
\midrule\midrule

SO-GAAL$^\dagger$
&0.545&0.435&0.877&\textbf{0.493}&0.107&\textbf{0.927}\\
\midrule

deepSVDD$^\dagger$
&\textbf{0.870}&\textbf{0.717}&\underline{0.942}&0.345&0.100&0.823\\
\midrule

LUNAR$^\dagger$

&0.490&0.294&0.809&0.282&0.093&0.794\\
\midrule

ICL$^\dagger$ %INTERNAL Contrastive Learning

&0.523&0.273&0.819&0.225&0.088&0.775\\
\midrule

BERT$^\dagger$
&\underline{0.861}&\underline{0.589}&\textbf{0.960}&0.281&0.082&0.784\\
\midrule

MLP$^\dagger$
&0.441&0.262&0.730&0.200&0.085&0.757\\
\midrule\midrule

MLP
&0.441&0.207&0.776&0.208&0.085&0.758\\
\midrule

MLP+Ours

& \makecell{0.639\\\scriptsize(+0.194)} 
& \makecell{0.404\\\scriptsize(+0.197)} 
& \makecell{0.841\\\scriptsize(+0.065)} 
& \makecell{\underline{0.424}\\\scriptsize(+0.216)} 
& \makecell{\textbf{0.259}\\\scriptsize(+0.173)} 
& \makecell{\underline{0.838}\\\scriptsize(+0.081)} \\

\bottomrule

\end{tabular}

\caption{
Performance on Anoshift benchmark. $\dagger$ denotes that metrics are reported from the results in the original paper. AUROC and AUPRC are denoted as ROC and PRC.
}
\label{table:anoshift}

\end{center}
\end{table}

%%%%%%%%%%%% Ablation %%%%%%%%%%%%
% Analysis on Our Method
% Each Component에 대한 분석
% MSL데이터, 잘 설명하기

% \newpage
\subsection{Ablation Study}
\label{ablation}

\begin{figure}[t!]
    \begin{center}
        \includegraphics[width=1.0\linewidth]{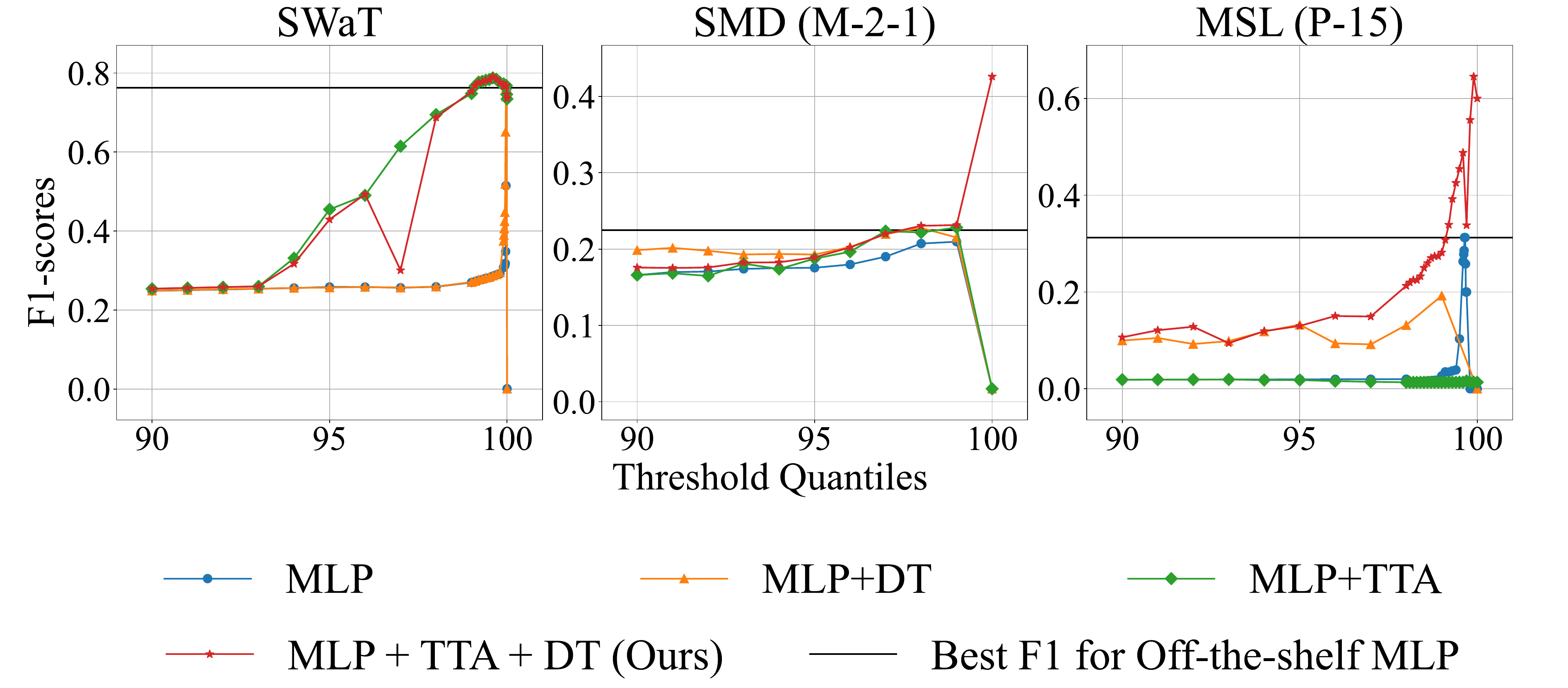}
    \end{center}
    \caption{
    F1 scores according to various thresholds.
    }
\label{fig:ab_f1trace}
\end{figure}

\begin{table*}[h!]
\begin{center}

\setlength{\tabcolsep}{5pt}
\small
\begin{tabular}[0.9\linewidth]{cc|cccc|cccc|cccc}
\toprule
\multirow{2}{*}{\textbf{DT}} & \multirow{2}{*}{\textbf{TTA}} 
& \multicolumn{4}{c|}{SWaT} 
& \multicolumn{4}{c|}{SMD (M-2-1)} 
& \multicolumn{4}{c}{MSL (P-15)} \\
\cmidrule{3-6} \cmidrule{7-10} \cmidrule{11-14}
 &  
 & F1 & F1-PA & AUROC & AUPRC 
 & F1 & F1-PA & AUROC & AUPRC 
 & F1 & F1-PA & AUROC & AUPRC \\
\midrule
\xmark & \xmark 
&0.765&0.834&0.832&0.722
&0.236&0.814&0.674&0.190
&0.263&0.848&0.645&0.061\\

\cmark & \xmark 
&0.762&0.837&0.846&0.738
&0.234&0.855&0.749&0.205
&0.221&0.703&0.799&0.124\\

\xmark & \cmark 
&\textbf{0.784}&\textbf{0.907}&0.888&0.778
&0.239&0.881&0.689&0.204
&0.019&0.027&0.640&0.060\\

\cmark & \cmark 
&\textbf{0.784}&0.903&\textbf{0.892}&\textbf{0.780}
&\textbf{0.249}&\textbf{0.974}&\textbf{0.764}&\textbf{0.280}
&\textbf{0.440}&\textbf{0.944}&\textbf{0.801}&\textbf{0.575}\\

\bottomrule
\end{tabular}

%\caption{Ablation study on our proposed method. \textbf{DT} and \textbf{TTA} indicate a detrend module and test-time adaptation, respectively.}
\caption{Ablation study on our proposed method. DT and TTA indicate a detrend module and test-time adaptation, respectively.}
\label{table:Ablation}

\end{center}
\end{table*}

% \begin{table}[h!]
% \begin{center}

% \caption{Ablation Study.}
% \label{table:Ablation}

% \setlength{\tabcolsep}{3pt}
% \begin{tabular}[1.0\linewidth]{l | l | ccccc}
% \toprule
% \footnotesize

% \thead{\textbf{Dataset}}
% & \thead{\textbf{Metrics}}
% & \thead{\textbf{MLP}}
% & \thead{\textbf{MLP}\\\textbf{+Detrend}}
% & \thead{\textbf{MLP}\\\textbf{+Adaptation}}
% & \thead{\textbf{MLP}\\\textbf{+Detrend}\\\textbf{+Adaptation}} \\

% \midrule\midrule

% \multirow{3}{*}{\thead{SWaT}}
% &F1&0.763&0.765&0.788&\textbf{0.791}\\
% &F1-PA&0.840&0.860&\textbf{0.919}&0.918\\
% &AUROC&0.828&0.843&0.887&\textbf{0.891}\\
% \midrule

% \multirow{3}{*}{\thead{SMD (M-2-1)}}
% &F1&0.225&0.228&0.240&\textbf{0.426}\\
% &F1-PA&0.841&0.871&\textbf{0.895}&0.892\\
% &AUROC&0.678&0.751&0.693&\textbf{0.842}\\
% \midrule

% \multirow{3}{*}{\thead{MSL (P-15)}}
% &F1&0.312&0.192&0.019&\textbf{0.645}\\
% &F1-PA&0.851&0.870&0.027&\textbf{0.976}\\
% &AUROC&0.647&\textbf{0.811}&0.546&0.809\\
% \midrule

% \multirow{3}{*}{\thead{CreditCard}}
% &F1&0.146&0.141&0.154&\textbf{0.162}\\
% &F1-PA&0.172&0.165&0.170&\textbf{0.173}\\
% &AUROC&0.946&0.939&\textbf{0.947}&0.946\\

% \bottomrule

% \end{tabular}
% \end{center}
% \vspace{-0.1in}
% \end{table}

As shown in Table~\ref{table:Ablation}, we perform the ablation study on our method to analyze the effectiveness of each component.
% Table~\ref{table:Ablation} shows the results with each component of our model. 
MLP with detrend module and test-time adaptation with the model update is consistently showing better results, compared to the cases when used alone (MLP+DT, MLP+TTA) and none of them used (MLP).
Here, DT and TTA denote a detrend module and test-time adaptation with model updates, respectively.
Moreover, Fig.~\ref{fig:ab_f1trace} also demonstrates that when the appropriate threshold is selected MLP model with our full method consistently outperforms these baselines, including the best performance of the off-the-shelf MLP model.

\begin{figure}[t!]
    \centering
    \includegraphics[width=1.0\linewidth]{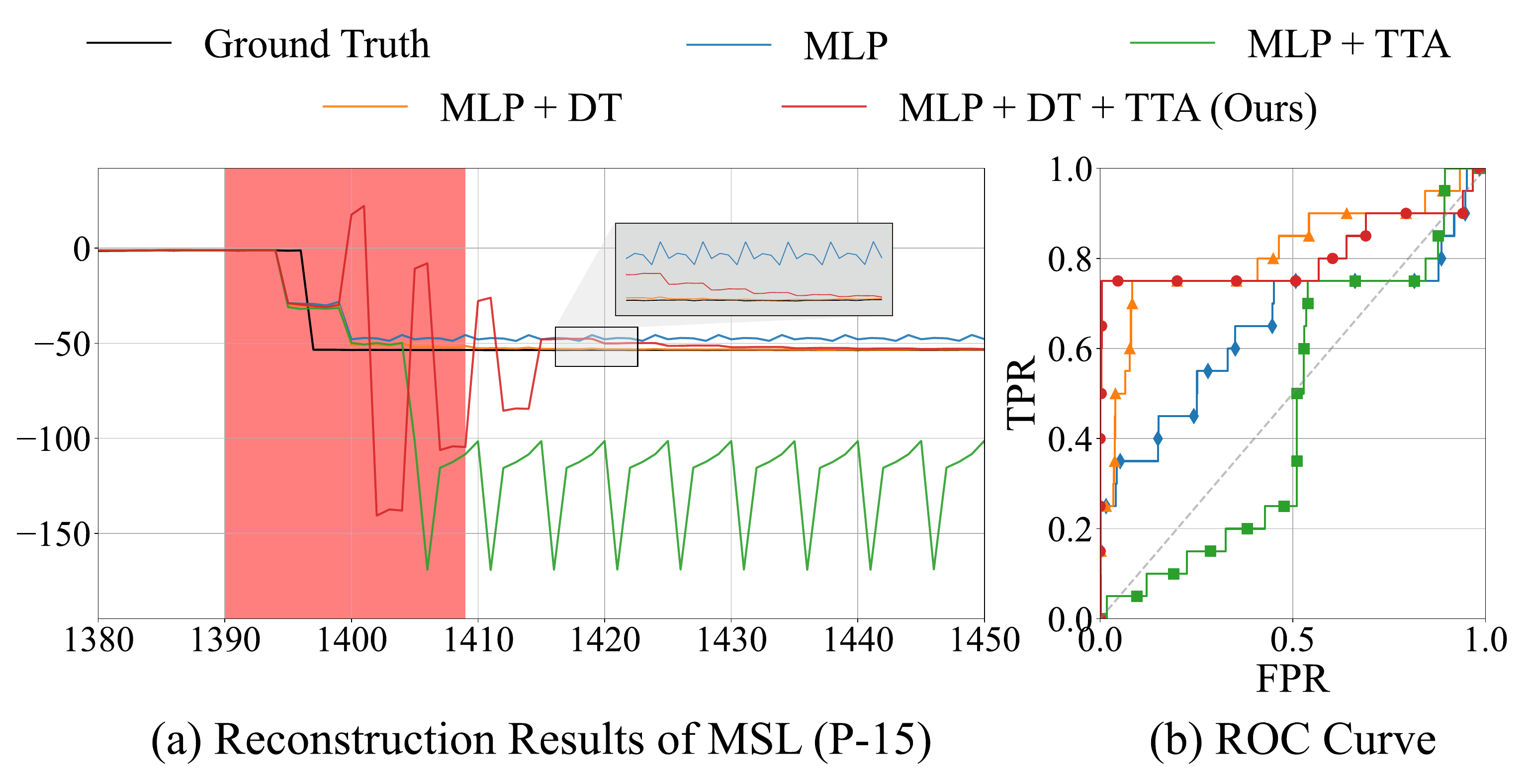}
    \caption{Ablation study on the proposed method using the MSL (P-15) dataset.}
\label{fig:ab_qual}
\end{figure}

This behavior can be further described in Fig.~\ref{fig:ab_qual}-(a), illustrating those four options at once. (1) Our approach (red) shows better reconstruction compared to off-the-shelf MLP (blue). The off-the-shelf MLP model is constantly generating reconstruction errors even after the transition of an overall trend, which results in many false positive cases. (2) Also, the detrend module alone fails to detect anomalies, showing less sensitivity compared to our approach, although they share the same EMA parameter $\gamma$. This shows model update can contribute to such sensitivity of the anomaly detector, as it keeps updating with recent observations. (3) Without proper update of such trend estimate, test-time adaptation with model updates alone (green) can harm the robustness of the model, as it can be overfitted to sequence before trend shift, with a lack of ability to adapt to newly coming sequences. 

\section{Discussion and Limitation}

\textbf{Threshold for Anomaly Detection.}
Existing unsupervised time-series anomaly detection studies~\cite{USAD, AnomalyTransformer} have a major limitation in that they determine the threshold for normality by inferring the entire test data and selecting it based on the best performance.
However, this approach is not practically feasible in real-world scenarios.
Therefore, we report AUROC to evaluate overall performance and decide the threshold based on the training data statistics in our experiments.
We posit that the performance of the anomaly detector could be further enhanced with an appropriate choice of threshold.
%We posit that the performance of test-time adaptation for time-series anomaly detection could be further enhanced if an appropriate threshold is determined for test-time adaptation.

% \noindent\textbf{Active Learning for Test-time Adaptation.}
% Our method updates the model by incorporating pseudo-labeling at test time.
% However, it should be noted that imprecise pseudo-labeling can occasionally have detrimental effects on the model's performance.
% Active learning~\cite{ren2021survey} improves the performances wherein human annotators provide labels for a subset of test data.
% Introducing user labeling for a subset of samples in test scenarios, aiming to enhance performance by test-time adaptation, represents a promising research direction in mitigating distribution shifts within anomaly detection.

\noindent\textbf{Inconsistent Labeling in Anomaly Detection.}
In the time-series anomaly detection task, the criteria of anomaly vary for each scenario, making it difficult to establish consistent labels.
For this reason, distinguishing whether test samples with significant differences from the normal in train sets are abnormal or normal with distribution shifts is challenging.
In our case, based on the assumption that there are more normal instances in test sets, we employ trend estimation and model predictions for test-time adaptation.
To improve the adaptation performance, employing active learning~\cite{ren2021survey} where human annotators provide labels for a subset of test data can be a valuable research direction.

\section{Conclusion}
In this work, we highlighted the distribution shift problem in unsupervised time-series anomaly detection. We have shown that the concept of normality may change over time. This can be a significant challenge for designing robust time-series anomaly detection frameworks, leading to many false positives, which harms the system's consistency. To mitigate this issue, we propose a simple yet effective strategy of incorporating new normals into the model architecture, by following trend estimates along with test-time adaptation. Concretely, our method consistently outperforms standard baselines for real-world benchmarks with such problems.

\section{Acknowledgements}
This work was supported by the Institute of Information \& communications Technology Planning \& Evaluation (IITP) grant funded by the Korea government (MSIT) (No.2019-0-00075, Artificial Intelligence Graduate School Program (KAIST))
and by the National Research Foundation of Korea (NRF) grant funded by the Korea government (MSIT) (No. NRF-2022R1A2B5B02001913 \& No. 2022R1A5A708390812).

% ============== References ============== %

\bibliography{references}

% ============== Appendix ============== %

\clearpage
\appendix

\section{Detailed Dataset Description}
\label{appendix:DD}

% \begin{description}[leftmargin=*]

\noindent\textbf{Secure Water Treatment (SWaT) and WAter DIstribution (WADI)~\cite{SWaT}.} 
% https://itrust.sutd.edu.sg/itrust-labs-home/itrust-labs_swat/
The SWaT and WADI\footnote{iTrust, Centre for Research in Cyber Security, Singapore University of Technology and Design} datasets encompass data collected from a testbed designed for cyber-physical systems involved in controlling water treatment processes. The SWaT dataset captures time-series data from 51 sensors and actuators deployed in industrial water treatment systems. This dataset spans a total of 11 days, consisting of 7 days of normal operation and 4 days for attack simulations. On the other hand, the WADI dataset focuses on a comprehensive system comprising water treatment, storage, and distribution networks. It includes 123 sensors deployed throughout the system. The dataset covers 16 days, encompassing 14 days of normal operations and 2 days of attack scenarios.

\noindent\textbf{Server Machine Dataset (SMD)~\cite{OmniAnomaly}.}
% https://github.com/NetManAIOps/OmniAnomaly
The SMD\footnote{https://github.com/NetManAIOps/OmniAnomaly} dataset, sourced from OmniAnomaly~\cite{OmniAnomaly}, consists of a comprehensive collection of data spanning five weeks and encompassing 28 distinct machines. To facilitate model training and evaluation, each observation from a machine is partitioned into two segments. The first half of the sequence serves as the training data, while the latter half is designated as the test data. we specifically selected two datasets, namely SMD-machine-1-4 and SMD-machine-2-1, which are characterized by significant distributional shifts.

\noindent\textbf{SMAP (Soil Moisture Active Passive) and Mars Science Laboratory (MSL).~\cite{HundmanCLCS18}}
The SMAP/MSL\footnote{https://github.com/khundman/telemanom} dataset comprises expert-labeled data extracted from the reports of spacecraft monitoring systems. Specifically, the SMAP dataset consists of 55 telemetry channels, while the MSL dataset contains 27 telemetry channels. Within these datasets, there are 28 unique entities referred to as Incident Surprise Anomaly (ISA) Reports in SMAP and 19 in MSL. For our experimental analysis, we specifically selected two ISA entities, namely MSL (P-15) and SMAP (T-3), to investigate and evaluate their anomaly detection performance.

\noindent\textbf{CreditCard.}
% https://www.kaggle.com/datasets/mlg-ulb/creditcardfraud
The CreditCard\footnote{https://www.kaggle.com/datasets/mlg-ulb/creditcardfraud} dataset consists of transaction logs of European cardholders spanning two days. For confidentiality reasons, the dataset only provides 28 principal component analysis (PCA) features, which have been anonymized, along with timestamp and transaction amount. The dataset is divided into two segments: the first half serves as the training dataset, while the latter half is designated as the test dataset.

\noindent\textbf{Yahoo.}
The Yahoo\footnote{https://webscope.sandbox.yahoo.com/} dataset comprises both real (A1) and synthetic (A2, A3, A4) time-series data for anomaly detection. Our primary focus was on the real-world data (A1) characterized by distribution shifts. From this subset, we specifically selected two datasets, namely A1-R20 and A1-R55, which both are univariate time-series. In our experimental setup, the initial 400 timesteps were used as the training dataset, while the remaining timesteps were allocated for testing purposes.

\begin{table*}[t!]
\begin{center}

\footnotesize
\setlength{\tabcolsep}{4pt}
\begin{tabular}[0.9\linewidth]{l | ccc ccc ccc}
% \begin{tabular}[1.0\linewidth]{>{\centering} l | p{1cm}p{1cm}p{1cm} p{1cm}p{1cm}p{1cm} p{1cm}p{1cm}p{1cm}}

\toprule

Dataset 
& \thead{SWaT} 
& \thead{WADI} 
& \thead{SMD\\(M-1-4)} 
& \thead{SMD\\(M-2-1)} 
& \thead{MSL\\(P-15)} 
& \thead{SMAP\\(T-3)} 
& \thead{CreditCard} 
& \thead{Yahoo\\(A1-R20)} 
& \thead{Yahoo\\(A1-R55)} \\
\toprule

$|\mathcal{D}_{train}|$
& 496800 & 784571 & 23706 & 23693 & 3682 & 2876 & 142403 & 400 & 400\\

$|\mathcal{D}_{test}|$
& 449919 & 172803 & 23707 & 23694 & 2856 & 8579 & 142404 & 1022 & 1027\\

$F$
& 51 & 123 & 38 & 38 & 55 & 25 & 29 & 1 & 1\\

$\frac{|\mathcal{D}_{test} \cap \mathcal{D}_{\pi} |}{|\mathcal{D}_{test}|}$
& 12.1\% & 5.77\% & 0.0304\% & 0.0494 & 0.00700 & 0.0212 & 0.00157 &0.0323 & 0.00487 \\

\bottomrule
\end{tabular}

\caption{
Dataset Descriptions. The number of observations in the training dataset and test dataset are denoted by $|\mathcal{D}_{train}|$ and $|\mathcal{D}_{test}|$ respectively. $F$ represents the number of features. Anomaly ratio during test-time is denoted as $\frac{|\mathcal{D}_{test} \cap \mathcal{D}_{\pi} |}{|\mathcal{D}_{test}|}$.
}

\label{table:Data_Description}

\end{center}
\end{table*}

% \end{description}

\section{Visualization Results of Datasets}
\label{appendix:TSNE}

\begin{figure*}[h]
    \centering    
    \includegraphics[width=1.0\linewidth]
    {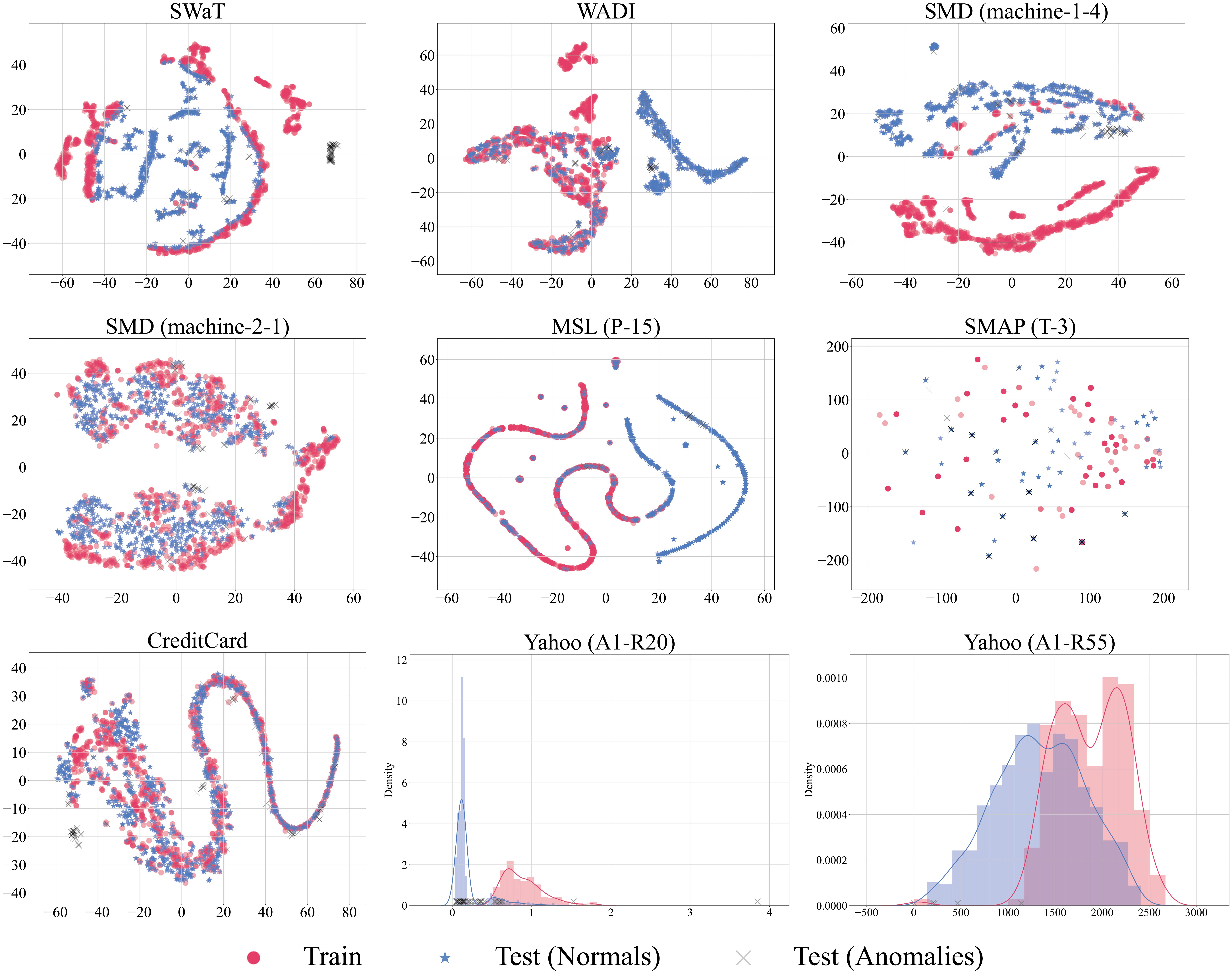}
    
    \caption{
        Visualization results of datasets. Red and blue each represent normal observations from the train and test datasets. Test-time anomalies are indicated by the symbol "X".
    }
    \label{fig:A_vis}
\end{figure*}

Figure~\ref{fig:A_vis} presents visualizations of the training and test dataset distributions for various benchmark datasets. T-SNE is used for visualization in all cases except for the Yahoo dataset, which contains univariate series.

It is worth noting that all datasets exhibit a "new normal problem", except for the CreditCard dataset.
These visualizations highlight distinct behavior between the training (red) and test (blue) datasets. 
This discrepancy implies that off-the-shelf models are prone to generating numerous false positives, compromising the reliability of the monitoring system.

Specifically, our approach addresses this issue in datasets such as SWaT and SMD (M-2-1), where test-time anomalies are distinguishable from both training and test distributions. By adapting to test-time normals, our approach enhances detection performance. 
In contrast, datasets such as SMD (M-1-4), MSL(P-15), and Yahoo exhibit test-time anomalies that are visually indistinguishable from test-time normals.

These are cases of contextual anomalies, where values remain within the range of normal behavior, but anomalous events are defined by significant deviations from recent context. 
Nevertheless, even in these cases, fitting the model to normal behavior significantly reduces false positives.

In the case of the WADI dataset, a trade-off emerges between reducing false positives and potentially missing true positives. 
Our approach reduces false positives by 73\% (from 102,824 to 75,076) while increasing true positives by 28\% (from 8,122 to 3,248) in WADI, as detailed in Table~\ref{table:Appendix:DetailExpAb}. 
The significance of balancing this trade-off depends on the specific application, as a decrease in true positives or an increase in false positives may have varying consequences. 
Exploring strategies to achieve this balance depending on the application is a promising avenue for future research. 
\section{Pseudocode for the proposed framework}
\label{appendix:TSNE}
In the method section, we suggest a simple algorithm for the model to adapt to new normals. In practice, our proposed methodology can be easily implemented by introducing an additional model update for the testing pipeline.
Anomaly detector is denoted as $f_{\theta}$ where $\theta$ being model parameters, initially trained with $\mathcal{D}_{train}$.  $\mathcal{D}_{test}$ with a stream of input length (or sliding window length) $w$ is assumed to arrive one after another. Hyperparameters include threshold $\tau$, test-time learning rate $\eta$, EMA update rate $\gamma$. Anomaly score of the stream is denoted as $\mathcal{A}(\mathcal{X}_{w,t})$.
 
\begin{algorithm}[H]
	\caption{Pseudocode for the proposed framework}
	\textbf{Input:} $f_{\theta}$, $\mathcal{D}_{test}$, $\tau$, $\eta$, $\gamma$.\\
    \textbf{Output: $\mathcal{A}(\mathcal{X}_{w,t})$} 
	\begin{algorithmic}[1]{
            \For {$\mathcal{X}_{w,t}\ in\  \mathcal{D}_{test}$} \Comment{Stream input of length $w$}
	        \State $\mu_{t} \leftarrow \gamma\mu_{t-w} + (1-\gamma)\hat{\mu} $ \Comment{Update trend estimate}
                \State $\mathcal{A}(\mathcal{X}_{w,t}) \leftarrow f_{\theta}(\mathcal{X}_{w,t}, \tau)$ \Comment{Anomaly Score of stream}
                \State $\hat{\mathcal{Y}}_{w, t} \leftarrow \mathcal{A}(\mathcal{X}_{w,t}) > \tau $ \Comment{Mask that filters anomalies}
                \State $\mathcal{L} (\cdot) \leftarrow  (1 - \hat{\mathcal{Y}}_{w,t})^\top \mathcal{L}(\mathcal{X}_{w,t})$ \Comment{Filter normals}
                \State $\theta \leftarrow \theta - \eta\nabla_{\theta}\mathcal{L}(\cdot)$ \Comment{Update model parameters}
	      \EndFor
       }
        \end{algorithmic} 
\label{alg:M2N2}
\end{algorithm}

\section{Computational Cost Analysis}
\label{appendix:CC}
\begin{table*}[t!]
\begin{center}

% \label{table:CC}

\footnotesize
\setlength{\tabcolsep}{3pt}
\begin{tabular}[1.0\linewidth]{l | ccccc}
\toprule

Measures 
& \thead{MLP}
& \thead{LSTM}
& \thead{USAD}
& \thead{THOC}
& \thead{AT}
\\

\midrule\midrule

Running Time (ms)
& \makecell{0.402\\\scriptsize($\pm$0.030)} 
& \makecell{5.674\\\scriptsize($\pm$0.823)} 
& \makecell{0.732\\\scriptsize($\pm$0.009)}
& \makecell{4.686\\\scriptsize($\pm$0.336)}
& \makecell{4.407\\\scriptsize($\pm$0.112)} \\

Total Flops
& 586740 & 84480 & 880110 & 1724360 & 4078068  \\

Parameter Size (MB)
& 2.35 & 2.90 & 3.53 & 1.82 & 1.32  \\

Forward/Backward Pass Size (MB)
& 0.01 & 0.03 & 0.02 & 0.11 & 0.33  \\

\bottomrule
\end{tabular}

\caption{Computational Cost of Baselines.}
\label{table:Appendix:CC}

\end{center}
\end{table*}
Table~\ref{table:Appendix:CC} presents a comprehensive analysis of the computational costs associated with the baseline methods. In order to ensure a fair and accurate comparison, the following parameters were kept constant: window size ($w$) was set to 12, the number of features ($F$) was set to 55, the hidden dimension ($d$) was set to 128, and the batch size was set to 1. The computational cost was measured using a Titan Xp with CUDA version 11.1.

To evaluate the running time, the mean of 30 trials was calculated, along with their standard deviation, for a single batch. MLP demonstrated the lowest computational time among all the baselines, whereas LSTM had the least total flops for the operation with the longest computational time.

% It is important to note that test-time adaptation requires additional memory cost, specifically for the forward and backward passes. MLP, in particular, incurs a memory cost that is less than 0.5\% of the parameter size, which makes test-time adaptation feasible with a reasonable cost.
% Note that test-time adaptation algorithms require additional memory cost, specifically for the forward and backward passes. 
% The inference times for our method and baselines are as follows: MLP ($0.402ms$), AT ($4.407ms$), THOC ($4.686ms$), and MLP+ours (including backward costs, $3.924ms$). 
% These results reveal that despite incurring backward costs, our method's inference time is lower than that of other complex baselines.
% Even more, the time required for prediction is practically similar to that of models without TTA, as the model update is performed after prediction is completed in TTA algorithms~\cite{TENT}.
% Therefore, we believe that our method exhibits practical applicability in real-world scenarios.

\section{Hyperparameter Settings}
\label{appendix:HypSet}

Table~\ref{table:Experiment_Settings},~\ref{table:Experiment_Settings_LSTM},~\ref{table:Experiment_Settings_USAD},~\ref{table:Experiment_Settings_THOC},~\ref{table:Experiment_Settings_AT} present the default hyperparameter settings for the baseline models, along with the test-time hyperparameters ($\eta$, $\gamma$) for the MLP model. 
The default hyperparameters specified in the respective papers for Anomaly Transformer~\cite{AnomalyTransformer}, USAD~\cite{USAD} and THOC~\cite{THOC} are used for training. 
The official implementation of Anomaly Transformer~\footnote{https://github.com/thuml/Anomaly-Transformer} is employed, while the other models are re-implemented based on the architectural specifications provided in the papers.

In the context of test-time adaptation, a lower learning rate ($\eta$) and update speed ($\gamma$) are favored for SWaT, WADI, and CreditCard datasets, where longer sequences of instances are available in the test data. Conversely, for other datasets with shorter test datasets such as SMD, MSL, SMAP, and Yahoo, were updated relatively faster in both learning rate and trend estimate. In order to update with an identical scheme through time, SGD without momentum is used for test-time parameter update.

\begin{table*}[t!]
\begin{center}

\footnotesize
\setlength{\tabcolsep}{3pt}
\begin{tabular}[1.0\linewidth]{l | ccccccccc}
\toprule

Datasets 
& \thead{SWaT}
& \thead{WADI}
& \thead{SMD\\(M-1-4)}
& \thead{SMD\\(M-2-1)}
& \thead{MSL\\(P-15)}
& \thead{SMAP\\(T-3)}
& \thead{CreditCard}
& \thead{Yahoo\\(A1-R20)}
& \thead{Yahoo\\(A1-R55)}
\\

\midrule\midrule

$W$
& 12 & 12 & 5 & 5 & 5 & 5 & 12 & 5 & 5  \\

$S_{train}$
& 12 & 12 & 5 & 5 & 5 & 5 & 12 & 1 & 1  \\

$S_{test}$
& 12 & 12 & 5 & 5 & 5 & 5 & 12 & 5 & 5 \\

$d$
& 128 & 128 & 16 & 16 & 16 & 16 & 128 & 2 & 2 \\

$\gamma$
& 0.99999 & 0.99 & 0.4 & 0.99 & 0.6 & 0.8 & 0.999 & 0.9 & 0.1 \\

$\eta$ 
& 0.005 & 0.001 & 0.1 & 0.05 & 0.1 & 1.0 & 0.01 & 0.005 & 0.005 \\

\bottomrule
\end{tabular}

\caption{Hyperparameters for MLP model.}
\label{table:Experiment_Settings}

\end{center}
\end{table*}

\begin{table*}[t!]
\begin{center}

\footnotesize
\setlength{\tabcolsep}{3pt}
\begin{tabular}[1.0\linewidth]{l | ccccccccc}
\toprule

Datasets 
& \thead{SWaT}
& \thead{WADI}
& \thead{SMD\\(M-1-4)}
& \thead{SMD\\(M-2-1)}
& \thead{MSL\\(P-15)}
& \thead{SMAP\\(T-3)}
& \thead{CreditCard}
& \thead{Yahoo\\(A1-R20)}
& \thead{Yahoo\\(A1-R55)}
\\

\midrule\midrule

$W$
& 12 & 12 & 5 & 5 & 5 & 5 & 12 & 5 & 5  \\

$S_{train}$
& 12 & 12 & 5 & 5 & 5 & 5 & 12 & 5 & 5  \\

$S_{test}$
& 12 & 12 & 5 & 5 & 5 & 5 & 12 & 5 & 5 \\

$d$
& 128 & 128 & 128 & 128 & 128 & 128 & 128 & 128 & 128 \\

\bottomrule
\end{tabular}

\caption{Hyperparameters for LSTM model.}
\label{table:Experiment_Settings_LSTM}

\end{center}
\end{table*}
\begin{table*}[t!]
\begin{center}

\footnotesize
\setlength{\tabcolsep}{3pt}
\begin{tabular}[1.0\linewidth]{l | ccccccccc}
\toprule

Datasets 
& \thead{SWaT}
& \thead{WADI}
& \thead{SMD\\(M-1-4)}
& \thead{SMD\\(M-2-1)}
& \thead{MSL\\(P-15)}
& \thead{SMAP\\(T-3)}
& \thead{CreditCard}
& \thead{Yahoo\\(A1-R20)}
& \thead{Yahoo\\(A1-R55)}
\\

\midrule\midrule

$W$
& 12 & 10 & 5 & 5 & 5 & 5 & 12 & 5 & 5  \\

$S_{train}$
& 12 & 10 & 5 & 5 & 5 & 5 & 12 & 1 & 1  \\

$S_{test}$
& 12 & 10 & 5 & 5 & 5 & 5 & 12 & 5 & 5 \\

$d$
& 40 & 40 & 38 & 38 & 33 & 55 & 40 & 40 & 40 \\

\bottomrule
\end{tabular}

\caption{Hyperparameters for USAD model.}
\label{table:Experiment_Settings_USAD}

\end{center}
\end{table*}
\begin{table*}[t!]
\begin{center}

\footnotesize
\setlength{\tabcolsep}{3pt}
\begin{tabular}[1.0\linewidth]{l | ccccccccc}
\toprule

Datasets 
& \thead{SWaT}
& \thead{WADI}
& \thead{SMD\\(M-1-4)}
& \thead{SMD\\(M-2-1)}
& \thead{MSL\\(P-15)}
& \thead{SMAP\\(T-3)}
& \thead{CreditCard}
& \thead{Yahoo\\(A1-R20)}
& \thead{Yahoo\\(A1-R55)}
\\

\midrule\midrule

$W$
& 100 & 100 & 100 & 100 & 100 & 100 & 100 & 100 & 100  \\

$S_{train}$
& 1 & 1 & 1 & 1 & 1 & 1 & 1 & 1 & 1  \\

$S_{test}$
& 1 & 1 & 1 & 1 & 1 & 1 & 1 & 1 & 1 \\

$d$
& 64 & 64 & 64 & 64 & 64 & 64 & 64 & 64 & 64 \\

\bottomrule
\end{tabular}

\caption{Hyperparameters for THOC model.}
\label{table:Experiment_Settings_THOC}

\end{center}
\end{table*}
\begin{table*}[t!]
\begin{center}

\footnotesize
\setlength{\tabcolsep}{3pt}
\begin{tabular}[1.0\linewidth]{l | ccccccccc}
\toprule

Datasets 
& \thead{SWaT}
& \thead{WADI}
& \thead{SMD\\(M-1-4)}
& \thead{SMD\\(M-2-1)}
& \thead{MSL\\(P-15)}
& \thead{SMAP\\(T-3)}
& \thead{CreditCard}
& \thead{Yahoo\\(A1-R20)}
& \thead{Yahoo\\(A1-R55)}
\\

\midrule\midrule

$W$
& 100 & 100 & 100 & 100 & 100 & 100 & 100 & 100 & 100  \\

$S_{train}$
& 100 & 100 & 100 & 100 & 100 & 100 & 100 & 100 & 100  \\

$S_{test}$
& 100 & 100 & 100 & 100 & 100 & 100 & 100 & 100 & 100  \\

$r$
& 0.5 & 0.5 & 0.5 & 0.5 & 1.0 & 1.0 & 0.5 & 1.0 & 1.0 \\

\bottomrule
\end{tabular}

\caption{Hyperparameters for Anomaly Transformer model.}
\label{table:Experiment_Settings_AT}

\end{center}
\end{table*}

\section{Detailed Experiment Results for Main Experiment and Ablation Study}
\label{appendix:DetailExp}
\begin{sidewaystable*}[]
\centering

\caption{Detailed results for baseline models, except for MLP.}
\label{table:Appendix:DetailExpMain}

\tiny
\resizebox{1.0\textheight}{!}{
\begin{tabular}{l | ccccc ccccc ccccc ccccc}
\toprule
&Metrics&Thr&$\tau$&Acc&Prec&Rec&F1&AUROC&TN&FP&FN&TP&Acc+&Prec+&Rec+&F1+&TN+&FP+&FN+&TP+\\

\midrule\midrule

\multirow{18}{*}{\rotatebox[origin=c]{90}{LSTMEncDec}}
&SWaT&$Q^*$&1.67E+07&0.827&0.377&0.646&0.476&0.749&337014&58321&19296&35288&0.848&0.434&0.819&0.567&337014&58321&9873&44711\\
&&Q100&1.98E+08&0.880&0.963&0.010&0.021&0.749&395313&22&54011&573&0.962&0.999&0.685&0.813&395313&22&17203&37381\\
&WADI&$Q^*$&9.67E+14&0.951&0.997&0.150&0.260&0.531&162821&5&8484&1493&0.952&0.997&0.162&0.279&162821&5&8360&1617\\
&&$Q^*$&9.67E+14&0.951&0.997&0.150&0.260&0.531&162821&5&8484&1493&0.952&0.997&0.162&0.279&162821&5&8360&1617\\
&SMD (M-1-4)&$Q^*$&4814.900&0.924&0.183&0.431&0.257&0.754&21604&1383&410&310&0.942&0.342&1&0.510&21604&1383&0&720\\
&&$Q^*$&4814.900&0.924&0.183&0.431&0.257&0.754&21604&1383&410&310&0.942&0.342&1&0.510&21604&1383&0&720\\
&SMD (M-2-1)&$Q^*$&1308.600&0.912&0.248&0.386&0.302&0.745&21151&1373&718&452&0.942&0.460&1&0.630&21151&1373&0&1170\\
&&Q100&812719.938&0.952&0.915&0.037&0.071&0.745&22520&4&1127&43&0.992&0.996&0.836&0.909&22520&4&192&978\\
&MSL (P-15)&$Q^*$&4.33E+07&0.952&0.040&0.250&0.068&0.599&2715&121&15&5&0.958&0.142&1&0.248&2715&121&0&20\\
&&$Q^*$&4.33E+07&0.952&0.040&0.250&0.068&0.599&2715&121&15&5&0.958&0.142&1&0.248&2715&121&0&20\\
&SMAP (T-3)&$Q^*$&6450.760&0.977&0.263&0.055&0.091&0.525&8369&28&172&10&0.997&0.867&1&0.929&8369&28&0&182\\
&&Q100&31647.510&0.978&0.222&0.011&0.021&0.525&8390&7&180&2&0.999&0.963&1&0.981&8390&7&0&182\\
&CreditCard&$Q^*$&10681.869&0.998&0.308&0.197&0.240&0.930&142082&99&179&44&0.998&0.336&0.224&0.269&142082&99&173&50\\
&&$Q^*$&10681.869&0.998&0.308&0.197&0.240&0.930&142082&99&179&44&0.998&0.336&0.224&0.269&142082&99&173&50\\
&Yahoo (A1-R20)&$Q^*$&0.052&0.635&0.036&0.394&0.065&0.408&636&353&20&13&0.655&0.085&1&0.158&636&353&0&33\\
&&Q99&0.186&0.795&0.027&0.152&0.046&0.408&808&181&28&5&0.816&0.126&0.788&0.217&808&181&7&26\\
&Yahoo (A1-R55)&$Q^*$&8.950&0.989&0.286&0.800&0.421&0.879&1012&10&1&4&0.989&0.286&0.800&0.421&1012&10&1&4\\
&&$Q^*$&8.950&0.989&0.286&0.800&0.421&0.879&1012&10&1&4&0.989&0.286&0.800&0.421&1012&10&1&4\\

\midrule

\multirow{18}{*}{\rotatebox[origin=c]{90}{USAD}}
&SWaT&$Q^*$&0.446&0.949&0.987&0.590&0.739&0.811&394910&425&22353&32231&0.959&0.988&0.668&0.797&394910&425&18106&36478\\
&&$Q^*$&0.446&0.949&0.987&0.590&0.739&0.811&394910&425&22353&32231&0.959&0.988&0.668&0.797&394910&425&18106&36478\\
&WADI&$Q^*$&585.845&0.951&0.997&0.150&0.260&0.543&162821&5&8484&1493&0.952&0.997&0.162&0.279&162821&5&8360&1617\\
&&$Q^*$&585.845&0.951&0.997&0.150&0.260&0.543&162821&5&8484&1493&0.952&0.997&0.162&0.279&162821&5&8360&1617\\
&SMD (M-1-4)&$Q^*$&20.129&0.969&0.397&0.037&0.069&0.374&22946&41&693&27&0.970&0.544&0.068&0.121&22946&41&671&49\\
&&Q96&0.309&0.942&0.057&0.058&0.058&0.374&22295&692&678&42&0.945&0.128&0.142&0.135&22295&692&618&102\\
&SMD (M-2-1)&$Q^*$&0.075&0.925&0.269&0.303&0.285&0.744&21561&963&815&355&0.959&0.549&1&0.708&21561&963&0&1170\\
&&Q98&0.113&0.951&0.519&0.115&0.189&0.744&22399&125&1035&135&0.991&0.897&0.928&0.912&22399&125&84&1086\\
&MSL (P-15)&$Q^*$&0.963&0.866&0.033&0.650&0.064&0.676&2460&376&7&13&0.868&0.051&1&0.096&2460&376&0&20\\
&&$Q^*$&0.963&0.866&0.033&0.650&0.064&0.676&2460&376&7&13&0.868&0.051&1&0.096&2460&376&0&20\\
&SMAP (T-3)&$Q^*$&0.020&0.441&0.023&0.615&0.045&0.520&3672&4725&70&112&0.449&0.037&1&0.072&3672&4725&0&182\\
&&Q99&0.121&0.976&0.071&0.011&0.019&0.520&8371&26&180&2&0.997&0.875&1&0.933&8371&26&0&182\\
&CreditCard&$Q^*$&0.017&0.998&0.292&0.359&0.322&0.931&141987&194&143&80&0.998&0.292&0.359&0.322&141987&194&143&80\\
&&$Q^*$&0.017&0.998&0.292&0.359&0.322&0.931&141987&194&143&80&0.998&0.292&0.359&0.322&141987&194&143&80\\
&Yahoo (A1-R20)&$Q^*$&0.852&0.960&0.167&0.061&0.089&0.389&979&10&31&2&0.983&0.722&0.788&0.754&979&10&7&26\\
&&$Q^*$&0.852&0.960&0.167&0.061&0.089&0.389&979&10&31&2&0.983&0.722&0.788&0.754&979&10&7&26\\
&Yahoo (A1-R55)&$Q^*$&0.462&0.996&1&0.200&0.333&0.875&1022&0&4&1&0.996&1&0.200&0.333&1022&0&4&1\\
&&Q99&0.290&0.988&0.182&0.400&0.250&0.875&1013&9&3&2&0.989&0.250&0.600&0.353&1013&9&2&3\\
\midrule

\multirow{18}{*}{\rotatebox[origin=c]{90}{THOC}}
&SWaT&$Q^*$&0.108&0.954&0.991&0.631&0.771&0.839&395008&327&20159&34425&0.966&0.992&0.726&0.838&395008&327&14974&39610\\
&&$Q^*$&0.108&0.954&0.991&0.631&0.771&0.839&395008&327&20159&34425&0.966&0.992&0.726&0.838&395008&327&14974&39610\\
&WADI&$Q^*$&0.099&0.267&0.070&0.947&0.130&0.494&36621&126205&530&9447&0.270&0.073&1&0.137&36621&126205&0&9977\\
&&Q97&0.100&0.440&0.061&0.600&0.110&0.494&70115&92711&3994&5983&0.463&0.097&1&0.177&70115&92711&0&9977\\
&SMD (M-1-4)&$Q^*$&0.101&0.927&0.207&0.490&0.291&0.812&21634&1353&367&353&0.940&0.323&0.897&0.475&21634&1353&74&646\\
&&$Q^*$&0.101&0.927&0.207&0.490&0.291&0.812&21634&1353&367&353&0.940&0.323&0.897&0.475&21634&1353&74&646\\
&SMD (M-2-1)&$Q^*$&0.102&0.885&0.214&0.497&0.300&0.682&20391&2133&588&582&0.910&0.354&1&0.523&20391&2133&0&1170\\
&&Q99&0.102&0.926&0.268&0.285&0.276&0.682&21613&911&837&333&0.962&0.562&1&0.720&21613&911&0&1170\\
&MSL (P-15)&$Q^*$&0.109&0.486&0.009&0.650&0.017&0.326&1374&1462&7&13&0.488&0.013&1&0.027&1374&1462&0&20\\
&&$Q^*$&0.109&0.486&0.009&0.650&0.017&0.326&1374&1462&7&13&0.488&0.013&1&0.027&1374&1462&0&20\\
&SMAP (T-3)&$Q^*$&0.100&0.967&0.220&0.214&0.217&0.611&8259&138&143&39&0.984&0.569&1&0.725&8259&138&0&182\\
&&Q97&0.102&0.972&0.191&0.099&0.130&0.611&8321&76&164&18&0.991&0.705&1&0.827&8321&76&0&182\\
&CreditCard&$Q^*$&0.104&0.998&0.192&0.184&0.188&0.864&142008&173&182&41&0.998&0.195&0.188&0.192&142008&173&181&42\\
&&$Q^*$&0.104&0.998&0.192&0.184&0.188&0.864&142008&173&182&41&0.998&0.195&0.188&0.192&142008&173&181&42\\
&Yahoo (A1-R20)&$Q^*$&0.102&0.170&0.037&1&0.072&0.157&141&848&0&33&0.170&0.037&1&0.072&141&848&0&33\\
&&$Q^*$&0.102&0.170&0.037&1&0.072&0.157&141&848&0&33&0.170&0.037&1&0.072&141&848&0&33\\
&Yahoo (A1-R55)&$Q^*$&0.107&0.884&0.025&0.600&0.048&0.847&905&117&2&3&0.884&0.025&0.600&0.048&905&117&2&3\\
&&$Q^*$&0.107&0.884&0.025&0.600&0.048&0.847&905&117&2&3&0.884&0.025&0.600&0.048&905&117&2&3\\
\midrule

\multirow{18}{*}{\rotatebox[origin=c]{90}{Anomaly Transformer}}
&SWaT&$Q^*$&0&0.121&0.121&1&0.216&0.382&0&395335&0&54584&0.121&0.121&1&0.216&0&395335&0&54584\\
&&Q99&0.003&0.861&0.081&0.014&0.024&0.382&386657&8678&53818&766&0.981&0.863&1&0.926&386657&8678&0&54584\\
&WADI&$Q^*$&0.000&0.672&0.072&0.396&0.122&0.540&112241&50585&6026&3951&0.707&0.165&1&0.283&112241&50585&0&9977\\
&&Q99&0.010&0.931&0.068&0.015&0.025&0.540&160763&2063&9827&150&0.988&0.829&1&0.906&160763&2063&0&9977\\
&SMD (M-1-4)&$Q^*$&0&0.030&0.030&1&0.059&0.508&0&22987&0&720&0.030&0.030&1&0.059&0&22987&0&720\\
&&Q99&0.091&0.960&0.045&0.015&0.023&0.508&22752&235&709&11&0.985&0.720&0.838&0.774&22752&235&117&603\\
&SMD (M-2-1)&$Q^*$&0.00E+00&0.049&0.049&1&0.094&0.500&0&22524&0&1170&0.049&0.049&1&0.094&0&22524&0&1170\\
&&Q99&0.170&0.942&0.059&0.012&0.020&0.500&22300&224&1156&14&0.985&0.823&0.893&0.857&22300&224&125&1045\\
&MSL (P-15)&$Q^*$&0.038&0.985&0.071&0.100&0.083&0.586&2810&26&18&2&0.991&0.435&1&0.606&2810&26&0&20\\
&&Q100&33.389&0.988&0.067&0.050&0.057&0.586&2822&14&19&1&0.995&0.588&1&0.741&2822&14&0&20\\
&SMAP (T-3)&$Q^*$&0.00E+00&0.021&0.021&1&0.042&0.460&0&8397&0&182&0.021&0.021&1&0.042&0&8397&0&182\\
&&Q99&0.003&0.969&0.022&0.011&0.015&0.460&8309&88&180&2&0.990&0.674&1&0.805&8309&88&0&182\\
&CreditCard&$Q^*$&3.800&0.998&0.155&0.067&0.094&0.549&142099&82&208&15&0.998&0.241&0.117&0.157&142099&82&197&26\\
&&$Q^*$&3.800&0.998&0.155&0.067&0.094&0.549&142099&82&208&15&0.998&0.241&0.117&0.157&142099&82&197&26\\
&Yahoo (A1-R20)&$Q^*$&0&0.032&0.032&1&0.063&0.493&0&989&0&33&0.032&0.032&1&0.063&0&989&0&33\\
&&Q98&0.000&0.929&0.024&0.030&0.027&0.493&948&41&32&1&0.934&0.146&0.212&0.173&948&41&26&7\\
&Yahoo (A1-R55)&$Q^*$&0.00E+00&0.005&0.005&1&0.010&0.483&0&1022&0&5&0.005&0.005&1&0.010&0&1022&0&5\\
&&$Q^*$&0.00E+00&0.005&0.005&1&0.010&0.483&0&1022&0&5&0.005&0.005&1&0.010&0&1022&0&5\\

\bottomrule
\end{tabular}
}
\end{sidewaystable*}
\begin{sidewaystable*}[]

\caption{Detailed results for ablation study.}
\label{table:Appendix:DetailExpAb}

\centering
\tiny
\resizebox{1.0\textheight}{!}{
\begin{tabular}{l | ccccc ccccc ccccc ccccc}
\toprule
&Metrics&Thr&$\tau$&Acc&Prec&Rec&F1&AUROC&TN&FP&FN&TP&Acc+&Prec+&Rec+&F1+&TN+&FP+&FN+&TP+\\

\midrule\midrule

\multirow{18}{*}{\rotatebox[origin=c]{90}{Off-the-shelf MLP}}
&SWaT&$Q^*$&6.954&0.953&0.988&0.621&0.763&0.828&394908&427&20667&33917&0.965&0.989&0.721&0.834&394908&427&15218&39366\\
&&Q99.98&4.723&0.949&0.918&0.633&0.750&0.828&392264&3071&20017&34567&0.965&0.931&0.765&0.840&392264&3071&12846&41738\\
&WADI&$Q^*$&0.141&0.394&0.073&0.814&0.134&0.499&60002&102824&1855&8122&0.405&0.088&1&0.163&60002&102824&0&9977\\
&&Q99.6&1.208&0.446&0.057&0.551&0.103&0.499&71629&91197&4480&5497&0.472&0.099&1&0.180&71629&91197&0&9977\\
&SMD (M-1-4)&$Q^*$&0.822&0.931&0.189&0.383&0.253&0.793&21801&1186&444&276&0.950&0.378&1&0.548&21801&1186&0&720\\
&&$Q^*$&0.822&0.931&0.189&0.383&0.253&0.793&21801&1186&444&276&0.950&0.378&1&0.548&21801&1186&0&720\\
&SMD (M-2-1)&$Q^*$&1.576&0.933&0.264&0.196&0.225&0.678&21885&639&941&229&0.973&0.647&1&0.785&21885&639&0&1170\\
&&Q99&1.882&0.940&0.299&0.162&0.210&0.678&22080&444&981&189&0.981&0.725&1&0.841&22080&444&0&1170\\
&MSL (P-15)&$Q^*$&11.173&0.992&0.417&0.250&0.312&0.647&2829&7&15&5&0.998&0.741&1&0.851&2829&7&0&20\\
&&$Q^*$&11.173&0.992&0.417&0.250&0.312&0.647&2829&7&15&5&0.998&0.741&1&0.851&2829&7&0&20\\
&SMAP (T-3)&$Q^*$&15.779&0.978&0.370&0.055&0.096&0.523&8380&17&172&10&0.998&0.915&1&0.955&8380&17&0&182\\
&&Q100&46.906&0.978&0.286&0.011&0.021&0.523&8392&5&180&2&0.999&0.973&1&0.986&8392&5&0&182\\
&CreditCard&$Q^*$&22.099&0.997&0.150&0.143&0.146&0.946&141999&182&191&32&0.997&0.173&0.170&0.172&141999&182&185&38\\
&&$Q^*$&22.099&0.997&0.150&0.143&0.146&0.946&141999&182&191&32&0.997&0.173&0.170&0.172&141999&182&185&38\\
&Yahoo (A1-R20)&$Q^*$&10.701&0.947&0.080&0.061&0.069&0.369&966&23&31&2&0.971&0.531&0.788&0.634&966&23&7&26\\
&&$Q^*$&10.701&0.947&0.080&0.061&0.069&0.369&966&23&31&2&0.971&0.531&0.788&0.634&966&23&7&26\\
&Yahoo (A1-R55)&$Q^*$&18.927&0.989&0.250&0.600&0.353&0.921&1013&9&2&3&0.990&0.308&0.800&0.444&1013&9&1&4\\
&&$Q^*$&18.927&0.989&0.250&0.600&0.353&0.921&1013&9&2&3&0.990&0.308&0.800&0.444&1013&9&1&4\\
\midrule

\multirow{18}{*}{\rotatebox[origin=c]{90}{MLP + DT}}
&SWaT&Q99.98&4.723&0.953&0.980&0.627&0.765&0.843&394622&713&20348&34236&0.965&0.982&0.721&0.832&394622&713&15218&39366\\
&&Q99.97&3.118&0.945&0.867&0.646&0.741&0.843&389935&5400&19315&35269&0.967&0.893&0.829&0.860&389935&5400&9353&45231\\
&WADI&$Q^*$&0.141&0.387&0.070&0.788&0.129&0.546&59045&103781&2116&7861&0.399&0.088&1&0.161&59045&103781&0&9977\\
&&Q99.7&1.447&0.541&0.058&0.456&0.103&0.546&88922&73904&5426&4551&0.572&0.119&1&0.213&88922&73904&0&9977\\
&SMD (M-1-4)&Q98&0.248&0.964&0.418&0.483&0.448&0.838&22503&484&372&348&0.980&0.598&1&0.748&22503&484&0&720\\
&&$Q^*$&0.822&0.972&0.568&0.260&0.357&0.838&22845&142&533&187&0.992&0.827&0.942&0.881&22845&142&42&678\\
&SMD (M-2-1)&Q98&1.136&0.927&0.237&0.219&0.228&0.751&21701&823&914&256&0.965&0.587&1&0.740&21701&823&0&1170\\
&&Q99&1.882&0.944&0.345&0.156&0.215&0.751&22176&348&987&183&0.985&0.771&1&0.871&22176&348&0&1170\\
&MSL (P-15)&Q99&3.461&0.985&0.156&0.250&0.192&0.811&2809&27&15&5&0.991&0.426&1&0.597&2809&27&0&20\\
&&$Q^*$&11.173&0.992&0.250&0.100&0.143&0.811&2830&6&18&2&0.998&0.769&1&0.870&2830&6&0&20\\
&SMAP (T-3)&Q95&0.007&0.945&0.097&0.192&0.129&0.581&8071&326&147&35&0.962&0.358&1&0.528&8071&326&0&182\\
&&Q100&46.906&0.979&0.333&0.011&0.021&0.581&8393&4&180&2&1.000&0.978&1&0.989&8393&4&0&182\\
&CreditCard&Q99.93&20.458&0.997&0.139&0.143&0.141&0.939&141982&199&191&32&0.997&0.160&0.170&0.165&141982&199&185&38\\
&&Q99.93&20.458&0.997&0.139&0.143&0.141&0.939&141982&199&191&32&0.997&0.160&0.170&0.165&141982&199&185&38\\
&Yahoo (A1-R20)&Q88&1.747&0.984&0.758&0.758&0.758&0.989&981&8&8&25&0.992&0.805&1&0.892&981&8&0&33\\
&&Q90&2.041&0.980&0.783&0.545&0.643&0.989&984&5&15&18&0.995&0.868&1&0.930&984&5&0&33\\
&Yahoo (A1-R55)&Q94&2.447&0.998&1&0.600&0.750&0.953&1022&0&2&3&0.999&1&0.800&0.889&1022&0&1&4\\
&&Q94&2.447&0.998&1&0.600&0.750&0.953&1022&0&2&3&0.999&1&0.800&0.889&1022&0&1&4\\

\midrule
    
\multirow{18}{*}{\rotatebox[origin=c]{90}{MLP + TTA}}
&SWaT&Q99.6&0.537&0.956&0.936&0.680&0.788&0.888&392795&2540&17472&37112&0.974&0.947&0.836&0.888&392795&2540&8977&45607\\
&&Q99.5&0.465&0.954&0.922&0.681&0.783&0.887&392212&3123&17418&37166&0.981&0.940&0.899&0.919&392212&3123&5515&49069\\
&WADI&Q92&0.090&0.390&0.074&0.832&0.136&0.492&59106&103720&1681&8296&0.400&0.088&1&0.161&59106&103720&0&9977\\
&&Q99.5&1.052&0.446&0.057&0.550&0.103&0.496&71657&91169&4486&5491&0.472&0.099&1&0.180&71657&91169&0&9977\\
&SMD (M-1-4)&$Q^*$&0.822&0.916&0.148&0.374&0.212&0.767&21443&1544&451&269&0.935&0.318&1&0.483&21443&1544&0&720\\
&&$Q^*$&0.822&0.916&0.148&0.374&0.212&0.767&21443&1544&451&269&0.935&0.318&1&0.483&21443&1544&0&720\\
&SMD (M-2-1)&$Q^*$&1.576&0.943&0.356&0.181&0.240&0.681&22140&384&958&212&0.984&0.753&1&0.859&22140&384&0&1170\\
&&Q99&1.882&0.947&0.404&0.159&0.228&0.693&22250&274&984&186&0.988&0.810&1&0.895&22250&274&0&1170\\
&MSL (P-15)&Q93&0.395&0.461&0.010&0.750&0.019&0.571&1301&1535&5&15&0.463&0.013&1&0.025&1301&1535&0&20\\
&&Q100&46.060&0.490&0.007&0.500&0.014&0.546&1390&1446&10&10&0.494&0.014&1&0.027&1390&1446&0&20\\
&SMAP (T-3)&$Q^*$&15.779&0.976&0.217&0.055&0.088&0.491&8361&36&172&10&0.996&0.835&1&0.910&8361&36&0&182\\
&&$Q^*$&15.779&0.976&0.217&0.055&0.088&0.491&8361&36&172&10&0.996&0.835&1&0.910&8361&36&0&182\\
&CreditCard&Q99.93&20.458&0.997&0.148&0.161&0.154&0.947&141973&208&187&36&0.997&0.161&0.179&0.170&141973&208&183&40\\
&&Q99.93&20.458&0.997&0.148&0.161&0.154&0.947&141973&208&187&36&0.997&0.161&0.179&0.170&141973&208&183&40\\
&Yahoo (A1-R20)&Q81&1.348&0.104&0.031&0.879&0.060&0.369&77&912&4&29&0.108&0.035&1&0.067&77&912&0&33\\
&&Q100&13.274&0.969&1&0.030&0.059&0.598&989&0&32&1&0.970&1&0.061&0.114&989&0&31&2\\
&Yahoo (A1-R55)&$Q^*$&18.927&0.997&1&0.400&0.571&0.910&1022&0&3&2&0.997&1&0.400&0.571&1022&0&3&2\\
&&$Q^*$&18.927&0.997&1&0.400&0.571&0.910&1022&0&3&2&0.997&1&0.400&0.571&1022&0&3&2\\
\midrule

\multirow{18}{*}{\rotatebox[origin=c]{90}{MLP + DT + TTA (Ours)}}
&SWaT&Q99.56
&0.626&0.955&0.954&0.664&0.783&0.889&393574&1761&18357&36227&0.978&0.963&0.848&0.902&393574&1761&8321&46263\\
&&Q99.3&0.399&0.953&0.905&0.682&0.778&0.891&391419&3916&17358&37226&0.980&0.927&0.909&0.918&391419&3916&4975&49609\\
&WADI&Q99.9&2.753&0.800&0.105&0.326&0.159&0.629&135078&27748&6729&3248&0.822&0.202&0.705&0.314&135078&27748&2943&7034\\
&&Q99.8&1.887&0.755&0.083&0.323&0.132&0.610&127243&35583&6754&3223&0.794&0.219&1&0.359&127243&35583&0&9977\\
&SMD (M-1-4)&Q97&0.215&0.968&0.469&0.490&0.480&0.845&22588&399&367&353&0.983&0.643&1&0.783&22588&399&0&720\\
&&Q99&0.384&0.971&0.545&0.356&0.430&0.845&22773&214&464&256&0.991&0.771&1&0.871&22773&214&0&720\\
&SMD (M-2-1)&Q99.76
&1.473&0.942&0.342&0.187&0.242&22102&422&951&219&0.982&0.735&1.0&0.847&22102&422&0&1170&0.703\\

&&Q99&1.882&0.947&0.403&0.162&0.231&0.713&22242&282&980&190&0.988&0.806&1&0.892&22242&282&0&1170\\
&MSL (P-15)&Q99.9&28.991&0.996&0.909&0.500&0.645&0.809&2835&1&10&10&1.000&0.952&1&0.976&2835&1&0&20\\
&&Q99.9&28.991&0.996&0.909&0.500&0.645&0.809&2835&1&10&10&1.000&0.952&1&0.976&2835&1&0&20\\
&SMAP (T-3)&Q96&0.078&0.958&0.153&0.220&0.180&0.605&8175&222&142&40&0.974&0.450&1.0&0.621&8175&222&0&182\\

&&Q98&0.971&0.973&0.152&0.055&0.081&0.625&8341&56&172&10&0.993&0.765&1&0.867&8341&56&0&182\\
&CreditCard&Q99.93&20.458&0.997&0.154&0.170&0.162&0.946&141972&209&185&38&0.997&0.164&0.184&0.173&141972&209&182&41\\
&&Q99.93&20.458&0.997&0.154&0.170&0.162&0.946&141972&209&185&38&0.997&0.164&0.184&0.173&141972&209&182&41\\
&Yahoo (A1-R20)&Q87&1.720&0.984&0.758&0.758&0.758&0.989&981&8&8&25&0.992&0.805&1&0.892&981&8&0&33\\
&&Q90&2.041&0.980&0.783&0.545&0.643&0.989&984&5&15&18&0.995&0.868&1&0.930&984&5&0&33\\
&Yahoo (A1-R55)&Q95&2.762&0.998&1&0.600&0.750&0.954&1022&0&2&3&0.999&1&0.800&0.889&1022&0&1&4\\
&&Q95&2.762&0.998&1&0.600&0.750&0.954&1022&0&2&3&0.999&1&0.800&0.889&1022&0&1&4\\

\bottomrule
\end{tabular}
}
\end{sidewaystable*}

Table~\ref{table:Appendix:DetailExpMain} and Table~\ref{table:Appendix:DetailExpAb} provide detailed results for the main experiment and ablation study, respectively. In these tables, various metrics are reported to evaluate the performance of the models. The "Thr" column represents the threshold applied, while the symbol $\tau$ denotes the exact threshold value. Additionally, $Qp$ represents the $p$-th percentile of the train anomaly score, and $Q^*$ represents the threshold that maximizes the F1-score using the test anomaly scores and test labels, which is calculated using the ROC curve.

The available metrics include Accuracy (ACC), Precision (Prec), Recall (Rec), F1-score (F1), True Negatives (TN), False Positives (FP), False Negatives (FN), and True Positives (TP). Metrics reported with a "+" sign indicate that a point-adjust is applied. The first and second lines of each entry in the tables represent detailed results that maximize the F1-score and F1-PA (F1 with Point Adjust), respectively. It should be noted that the off-the-shelf baselines utilize $Q^*$ as the threshold to have maximum achievable performance for off-the-shelf baselines, while our approach sets the threshold using only the train anomaly scores.
\label{appendix:Main_Experiment_Full}
\begin{table*}[t!]
\begin{center}

\setlength{\tabcolsep}{2.5pt}
\small
\begin{tabular}{@{}l|l|ccccc|c}
\toprule
\textbf{Dataset} & \textbf{Metrics} & \textbf{MLP} & \textbf{LSTM} & \textbf{USAD} & \textbf{THOC} & \textbf{AT} & \textbf{Ours} \\
\midrule\midrule

\multirowcell{4}{SWaT}
&F1&\ocell[0.765]{0.002}&\ocell[0.401]{0.060}&\ocell[0.557]{0.174}&\ocell[0.776]{0.012}&\ocell[0.218]{0.003}&\ocell[0.784]{0.003}\\
&F1-PA&\ocell[0.831]{0.004}&\ocell[0.768]{0.104}&\ocell[0.655]{0.182}&\ocell[0.862]{0.030}&\ocell[0.962]{0.002}&\ocell[0.903]{0.003}\\
&AUROC&\ocell[0.832]{0.003}&\ocell[0.697]{0.051}&\ocell[0.737]{0.110}&\ocell[0.838]{0.005}&\ocell[0.530]{0.020}&\ocell[0.892]{0.003}\\
&AUPRC&\ocell[0.722]{0.004}&\ocell[0.248]{0.046}&\ocell[0.457]{0.228}&\ocell[0.744]{0.008}&\ocell[0.195]{0.033}&\ocell[0.780]{0.002}\\

\midrule

\multirowcell{4}{WADI} 
&F1&\ocell[0.131]{0.002}&\ocell[0.245]{0.015}&\ocell[0.260]{0.000}&\ocell[0.124]{0.004}&\ocell[0.109]{0.000}&\ocell[0.148]{0.010}\\
&F1-PA&\ocell[0.175]{0.010}&\ocell[0.279]{0.000}&\ocell[0.279]{0.000}&\ocell[0.153]{0.013}&\ocell[0.915]{0.003}&\ocell[0.346]{0.022}\\
&AUROC&\ocell[0.485]{0.005}&\ocell[0.525]{0.004}&\ocell[0.530]{0.029}&\ocell[0.484]{0.008}&\ocell[0.501]{0.002}&\ocell[0.624]{0.004}\\
&AUPRC&\ocell[0.052]{0.000}&\ocell[0.195]{0.009}&\ocell[0.205]{0.003}&\ocell[0.144]{0.088}&\ocell[0.059]{0.004}&\ocell[0.081]{0.002}\\

\midrule

\multirowcell{4}{SMD\\(M-1-4)}
&F1&\ocell[0.273]{0.011}&\ocell[0.282]{0.009}&\ocell[0.159]{0.023}&\ocell[0.379]{0.026}&\ocell[0.059]{0.000}&\ocell[0.463]{0.006}\\
&F1-PA&\ocell[0.544]{0.026}&\ocell[0.500]{0.016}&\ocell[0.296]{0.043}&\ocell[0.521]{0.023}&\ocell[0.799]{0.020}&\ocell[0.874]{0.006}\\
&AUROC&\ocell[0.805]{0.007}&\ocell[0.818]{0.029}&\ocell[0.673]{0.044}&\ocell[0.869]{0.032}&\ocell[0.479]{0.011}&\ocell[0.845]{0.002}\\
&AUPRC&\ocell[0.169]{0.006}&\ocell[0.151]{0.008}&\ocell[0.103]{0.010}&\ocell[0.223]{0.019}&\ocell[0.034]{0.003}&\ocell[0.354]{0.005}\\

\midrule

\multirowcell{4}{SMD\\(M-2-1)}
&F1&\ocell[0.236]{0.008}&\ocell[0.283]{0.014}&\ocell[0.308]{0.051}&\ocell[0.295]{0.038}&\ocell[0.094]{0.000}&\ocell[0.249]{0.137}\\
&F1-PA&\ocell[0.814]{0.023}&\ocell[0.910]{0.010}&\ocell[0.922]{0.002}&\ocell[0.705]{0.038}&\ocell[0.866]{0.017}&\ocell[0.974]{0.004}\\
&AUROC&\ocell[0.674]{0.015}&\ocell[0.727]{0.007}&\ocell[0.738]{0.040}&\ocell[0.668]{0.031}&\ocell[0.498]{0.004}&\ocell[0.764]{0.022}\\
&AUPRC&\ocell[0.190]{0.008}&\ocell[0.251]{0.011}&\ocell[0.246]{0.049}&\ocell[0.161]{0.031}&\ocell[0.052]{0.006}&\ocell[0.280]{0.097}\\

\midrule

\multirowcell{4}{MSL\\(P-15)} 
&F1&\ocell[0.263]{0.061}&\ocell[0.056]{0.007}&\ocell[0.060]{0.005}&\ocell[0.018]{0.000}&\ocell[0.071]{0.038}&\ocell[0.440]{0.183}\\
&F1-PA&\ocell[0.848]{0.028}&\ocell[0.351]{0.146}&\ocell[0.097]{0.013}&\ocell[0.027]{0.000}&\ocell[0.437]{0.357}&\ocell[0.944]{0.018}\\
&AUROC&\ocell[0.645]{0.006}&\ocell[0.617]{0.011}&\ocell[0.661]{0.010}&\ocell[0.332]{0.004}&\ocell[0.568]{0.088}&\ocell[0.801]{0.011}\\
&AUPRC&\ocell[0.061]{0.017}&\ocell[0.012]{0.002}&\ocell[0.016]{0.001}&\ocell[0.005]{0.000}&\ocell[0.023]{0.015}&\ocell[0.575]{0.094}\\

\midrule

\multirowcell{4}{SMAP\\(T-3)}
&F1&\ocell[0.095]{0.002}&\ocell[0.091]{0.007}&\ocell[0.044]{0.001}&\ocell[0.154]{0.089}&\ocell[0.042]{0.000}&\ocell[0.218]{0.032}\\
&F1-PA&\ocell[0.992]{0.007}&\ocell[0.998]{0.003}&\ocell[0.940]{0.008}&\ocell[0.747]{0.074}&\ocell[0.772]{0.027}&\ocell[0.708]{0.045}\\
&AUROC&\ocell[0.510]{0.012}&\ocell[0.515]{0.012}&\ocell[0.500]{0.020}&\ocell[0.591]{0.062}&\ocell[0.490]{0.007}&\ocell[0.617]{0.007}\\
&AUPRC&\ocell[0.044]{0.003}&\ocell[0.050]{0.011}&\ocell[0.031]{0.001}&\ocell[0.049]{0.029}&\ocell[0.017]{0.003}&\ocell[0.111]{0.015}\\

\midrule

\multirowcell{4}{Credit\\Card} 
&F1&\ocell[0.127]{0.009}&\ocell[0.220]{0.009}&\ocell[0.323]{0.062}&\ocell[0.138]{0.045}&\ocell[0.039]{0.027}&\ocell[0.135]{0.011}\\
&F1-PA&\ocell[0.145]{0.015}&\ocell[0.234]{0.019}&\ocell[0.323]{0.062}&\ocell[0.148]{0.045}&\ocell[0.056]{0.041}&\ocell[0.151]{0.009}\\
&AUROC&\ocell[0.943]{0.003}&\ocell[0.930]{0.005}&\ocell[0.887]{0.048}&\ocell[0.770]{0.027}&\ocell[0.548]{0.033}&\ocell[0.943]{0.002}\\
&AUPRC&\ocell[0.055]{0.004}&\ocell[0.109]{0.008}&\ocell[0.234]{0.065}&\ocell[0.041]{0.018}&\ocell[0.007]{0.005}&\ocell[0.063]{0.003}\\

\midrule

\multirowcell{4}{Yahoo\\(A1-R20)} 
&F1&\ocell[0.067]{0.002}&\ocell[0.065]{0.001}&\ocell[0.277]{0.038}&\ocell[0.106]{0.017}&\ocell[0.098]{0.023}&\ocell[0.678]{0.170}\\
&F1-PA&\ocell[0.259]{0.243}&\ocell[0.426]{0.216}&\ocell[0.695]{0.086}&\ocell[0.106]{0.017}&\ocell[0.185]{0.042}&\ocell[0.895]{0.051}\\
&AUROC&\ocell[0.367]{0.001}&\ocell[0.394]{0.012}&\ocell[0.668]{0.031}&\ocell[0.198]{0.022}&\ocell[0.525]{0.017}&\ocell[0.971]{0.036}\\
&AUPRC&\ocell[0.056]{0.000}&\ocell[0.057]{0.001}&\ocell[0.161]{0.031}&\ocell[0.067]{0.024}&\ocell[0.048]{0.013}&\ocell[0.637]{0.153}\\

\midrule

\multirowcell{4}{Yahoo\\(A1-R55)} 
&F1&\ocell[0.366]{0.033}&\ocell[0.446]{0.029}&\ocell[0.281]{0.071}&\ocell[0.059]{0.010}&\ocell[0.010]{0.000}&\ocell[0.633]{0.235}\\
&F1-PA&\ocell[0.424]{0.070}&\ocell[0.446]{0.029}&\ocell[0.320]{0.093}&\ocell[0.059]{0.010}&\ocell[0.010]{0.000}&\ocell[0.744]{0.290}\\
&AUROC&\ocell[0.916]{0.006}&\ocell[0.877]{0.044}&\ocell[0.867]{0.022}&\ocell[0.875]{0.028}&\ocell[0.478]{0.004}&\ocell[0.958]{0.011}\\
&AUPRC&\ocell[0.303]{0.011}&\ocell[0.242]{0.075}&\ocell[0.177]{0.096}&\ocell[0.019]{0.004}&\ocell[0.002]{0.000}&\ocell[0.624]{0.004}\\

\bottomrule

\end{tabular}

\caption{
Confidence interval report for the main experiment. 
}
\label{table:Main_Exp_Full}

\end{center}
\end{table*}
\begin{table*}[h!]
\begin{center}

\setlength{\tabcolsep}{5pt}
\small
\begin{tabular}[0.9\linewidth]{cc|cccc|cccc|cccc}
\toprule
\multirow{2}{*}{\textbf{DT}} & \multirow{2}{*}{\textbf{TTA}} 
& \multicolumn{4}{c|}{SWaT} 
& \multicolumn{4}{c|}{SMD (M-2-1)} 
& \multicolumn{4}{c}{MSL (P-15)} \\
\cmidrule{3-6} \cmidrule{7-10} \cmidrule{11-14}
 &  
 & F1 & F1-PA & AUROC & AUPRC 
 & F1 & F1-PA & AUROC & AUPRC 
 & F1 & F1-PA & AUROC & AUPRC \\
\midrule
\xmark & \xmark 
&\cell[0.765]{0.002}&\cell[0.834]{0.003}&\cell[0.832]{0.003}&\cell[0.722]{0.004}
&\cell[0.236]{0.008}&\cell[0.814]{0.023}&\cell[0.674]{0.015}&\cell[0.190]{0.008}
&\cell[0.263]{0.061}&\cell[0.848]{0.028}&\cell[0.645]{0.006}&\cell[0.061]{0.017}\\

\cmark & \xmark 
&\cell[0.762]{0.006}&\cell[0.837]{0.014}&\cell[0.846]{0.005}&\cell[0.738]{0.004}
&\cell[0.234]{0.007}&\cell[0.855]{0.018}&\cell[0.749]{0.011}&\cell[0.205]{0.005}
&\cell[0.221]{0.019}&\cell[0.703]{0.352}&\cell[0.799]{0.008}&\cell[0.124]{0.015}\\

\xmark & \cmark 
&\cell[0.784]{0.001}&\cell[0.907]{0.007}&\cell[0.888]{0.003}&\cell[0.778]{0.002}
&\cell[0.239]{0.006}&\cell[0.881]{0.018}&\cell[0.689]{0.007}&\cell[0.204]{0.006}
&\cell[0.019]{0.000}&\cell[0.027]{0.000}&\cell[0.640]{0.036}&\cell[0.060]{0.026}\\

\cmark & \cmark 
&\cell[0.784]{0.003}&\cell[0.903]{0.003}&\cell[0.892]{0.003}&\cell[0.780]{0.002}
&\cell[0.249]{0.137}&\cell[0.974]{0.004}&\cell[0.764]{0.022}&\cell[0.280]{0.097}
&\cell[0.440]{0.183}&\cell[0.944]{0.018}&\cell[0.801]{0.011}&\cell[0.575]{0.094}\\

\bottomrule
\end{tabular}

\caption{
Confidence interval report for the ablation study.
}
\label{table:Ablation_Full}

\end{center}
\end{table*}

% \begin{table}[h!]
% \begin{center}

% \caption{Ablation Study.}
% \label{table:Ablation}

% \setlength{\tabcolsep}{3pt}
% \begin{tabular}[1.0\linewidth]{l | l | ccccc}
% \toprule
% \footnotesize

% \thead{\textbf{Dataset}}
% & \thead{\textbf{Metrics}}
% & \thead{\textbf{MLP}}
% & \thead{\textbf{MLP}\\\textbf{+Detrend}}
% & \thead{\textbf{MLP}\\\textbf{+Adaptation}}
% & \thead{\textbf{MLP}\\\textbf{+Detrend}\\\textbf{+Adaptation}} \\

% \midrule\midrule

% \multirow{3}{*}{\thead{SWaT}}
% &F1&0.763&0.765&0.788&\textbf{0.791}\\
% &F1-PA&0.840&0.860&\textbf{0.919}&0.918\\
% &AUROC&0.828&0.843&0.887&\textbf{0.891}\\
% \midrule

% \multirow{3}{*}{\thead{SMD (M-2-1)}}
% &F1&0.225&0.228&0.240&\textbf{0.426}\\
% &F1-PA&0.841&0.871&\textbf{0.895}&0.892\\
% &AUROC&0.678&0.751&0.693&\textbf{0.842}\\
% \midrule

% \multirow{3}{*}{\thead{MSL (P-15)}}
% &F1&0.312&0.192&0.019&\textbf{0.645}\\
% &F1-PA&0.851&0.870&0.027&\textbf{0.976}\\
% &AUROC&0.647&\textbf{0.811}&0.546&0.809\\
% \midrule

% \multirow{3}{*}{\thead{CreditCard}}
% &F1&0.146&0.141&0.154&\textbf{0.162}\\
% &F1-PA&0.172&0.165&0.170&\textbf{0.173}\\
% &AUROC&0.946&0.939&\textbf{0.947}&0.946\\

% \bottomrule

% \end{tabular}
% \end{center}
% \vspace{-0.1in}
% \end{table}
\section{Visualization of Anomaly Scores}
\label{appendix:Anosc}

\begin{figure*}[t!]
    \begin{center}
        \includegraphics[width=1.0\linewidth]{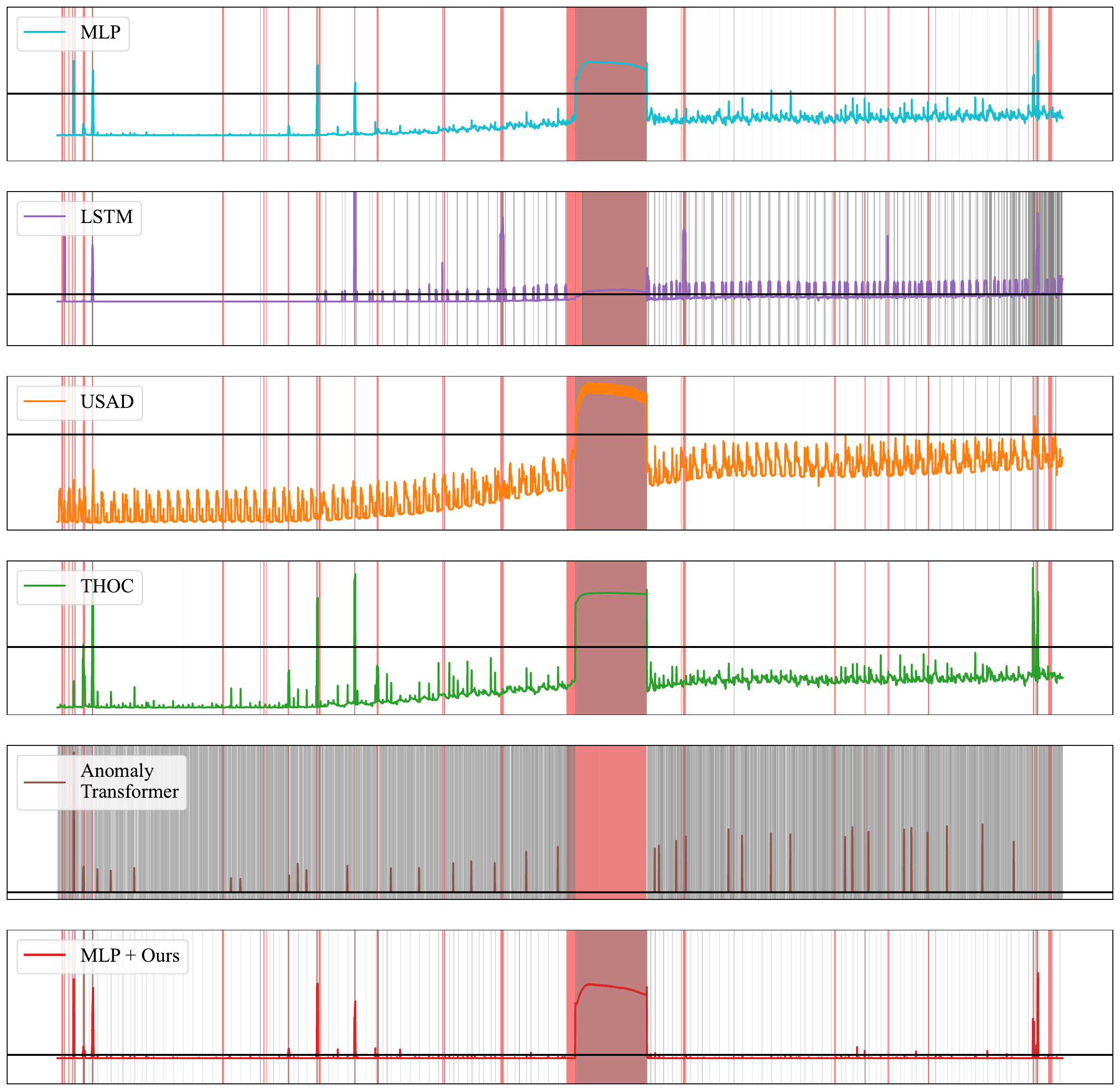}
    \end{center}
    \caption{
    Visualization of anomaly scores for various baselines and our method. The black horizontal line represents the threshold ($\tau$) applied, the red shade indicates the ground truth anomaly labels, and the gray shade represents the models' predictions of anomalies.
    }
\label{fig:A_anosc}
\end{figure*}

Figure~\ref{fig:A_anosc} presents the visualization results of anomaly scores, $\mathcal{A}(X_t)$, for the SWaT test dataset. For MLP, the anomaly score is defined as the channel-wise mean of reconstruction errors for each timestep. For other baselines, anomaly scores are calculated following the definition of $\mathcal{A}(X_t)$ for each model. 

The upper five rows of figures display the anomaly scores obtained from off-the-shelf baselines (MLP, LSTM, USAD, THOC, AT), while the lower figure corresponds to MLP with our adaptation strategy. The black horizontal line represents the threshold ($\tau$) applied, the red shade indicates the ground truth anomaly labels, and the gray shade represents the models' predictions of anomalies. To facilitate visualization, all anomaly scores, including thresholds, have been scaled to the range of 0 to 1.

It is evident from the visualization that all off-the-shelf baselines exhibit gradually increasing anomaly scores. This indicates that these models have failed to adapt to the changing trend and dynamics observed during the test phase. In contrast, our proposed approach demonstrates robustness in the face of such shifts, which serves as the primary driver behind the observed performance improvements.

\end{document}